\documentclass{midl} 

\usepackage{mwe} 

\newcommand{\ours}{CAPRMIL }

\usepackage{makecell}
\usepackage{booktabs}
\usepackage{multirow}
\usepackage{graphicx}
\usepackage[format=plain]{caption}
\jmlrproceedings{}{}
\jmlrpages{}{}

\title[]{CAPRMIL: Context-Aware Patch Representations for Multiple Instance Learning}

\midlauthor{\Name{Andreas Lolos\midljointauthortext{Contributed equally}\nametag{$^{1,2}$}} \orcid{0009-0000-6593-5999} \Email{andreaslolos@phys.uoa.gr}\\
\Name{Theofilos Christodoulou\midlotherjointauthor\nametag{$^{2}$}} \Email{th.christodoulou@athenarc.gr}\\
\Name{Aris L. Moustakas\nametag{$^{1,2}$}} \Email{arislm@phys.uoa.gr}\\
\Name{Stergios Christodoulidis\nametag{$^{3,4}$}} \Email{stergios.christodoulidis@centralesupelec.fr}\\
\Name{Maria Vakalopoulou\nametag{$^{2,3,4}$}} \Email{maria.vakalopoulou@centralesupelec.fr}\\
\addr $^{1}$ National and Kapodistrian University of Athens, Greece \\
\addr $^{2}$ Archimedes, Athena Research Center, Greece \\
\addr $^3$ MICS Laboratory, CentraleSupélec, Université Paris-Saclay \\
\addr $^4$ IHU PRISM, National Center for Precision Medicine in Oncology, Gustave Roussy \\
}

\begin{document}

\maketitle

\begin{abstract}
In computational pathology, weak supervision has become the standard for deep learning due to the gigapixel scale of WSIs and the scarcity of pixel-level annotations, with Multiple Instance Learning (MIL) established as the principal framework for slide-level model training.
In this paper, we introduce a novel setting for MIL methods, inspired by proceedings in Neural Partial Differential Equation (PDE) Solvers. Instead of relying on complex attention-based aggregation, we propose an efficient, aggregator-agnostic framework that removes the complexity of correlation learning from the MIL aggregator. \ours produces rich context-aware patch embeddings that promote effective correlation learning on downstream tasks. By projecting patch features -extracted using a frozen patch encoder- into a small set of global context/morphology-aware tokens and utilizing multi-head self-attention, \ours injects global context with linear computational complexity with respect to the bag size. Paired with a simple Mean MIL aggregator, \ours matches  
state-of-the-art slide-level performance across multiple public pathology benchmarks, while reducing the total number of trainable parameters by $48\%-92.8\%$ versus SOTA MILs, lowering FLOPs during inference by $52\%-99\%$, and ranking among the best models on GPU memory efficiency and training time. Our results indicate that learning rich, context-aware instance representations before aggregation is an effective and scalable alternative to complex pooling for whole-slide analysis. Our code is available at: \url{https://github.com/mandlos/CAPRMIL}

\end{abstract}


\section{Introduction}

Whole Slide Image (WSI) analysis has become the foundation of clinical practice in computational pathology \cite{alkhalaf_integration_2024, wang_advances_2024}, however their sheer size poses a significant challenge for Deep Learning approaches \cite{brixtel_whole_2022, lu_data-efficient_2021, gadermayr_multiple_2024}. At the same time, pixel-level annotations are prohibitively expensive and time-consuming, resulting in clinical datasets typically providing only slide-level labels rather than fine-grained annotations. \cite{lu_data-efficient_2021, song_artificial_2023, gadermayr_multiple_2024}.

To address the computationally prohibitive size of WSIs and the lack of pixel-level annotations, Multiple Instance Learning (MIL) has been established as the standard framework for WSI analysis. 
The MIL pipeline comprises patch feature extraction, typically adopting pre-trained foundation models \cite{xiong_survey_2025}, followed by aggregation/pooling to produce the slide-level representation for downstream tasks.
In recent years, attention-based mechanisms have emerged as a promising approach for a trainable MIL aggregator \cite{ilse_attention-based_2018, wang_advances_2024, gadermayr_multiple_2024}, due to their impressive correlation learning capabilities.
While effective, approaches that utilize standard attention directly on the patch embeddings face computational bottlenecks due to the quadratic complexity of the attention operator~\cite{shao_transmil_2021}. Attention-based MIL methods for WSI have also been found to be highly susceptible to overfitting and offer limited interpretability \cite{zhang_attention-challenging_2025}, while often lacking principled uncertainty quantification \cite{sun_prototype-based_2026, cui_bayes-mil_2022, lolos_sgpmil_2025}, limiting the potential of clinical translation. 
Therefore, developing aggregation strategies that can effectively model instance interactions, handle the challenges inherent to long sequence processing in WSIs, and provide reliable representations remains an active area of research \cite{bilal_aggregation_2023, fang_mammil_2024}.


At the same time, we identify that neural Partial Differential Equation (PDE) Solvers \cite{li_fourier_2020, hao_gnot_2023, wu_transolver_2024} face a similar challenge: how to achieve efficient and reliable correlation learning in large-scale inputs. Solving PDEs often includes modeling complex phenomena that may cause long-distance interactions, on domains discretized into millions of mesh points \cite{grossmann_can_2024}. 
Attention-based methods have been used in PDE modeling, but they also face
prohibitive computational cost and degraded correlation learning due to the large scale of the input \cite{katharopoulos_transformers_2020, wu_transolver_2024}. Therefore, we assume that ideas that have successfully tackled these problems in the domain of Surrogate PDE Solvers could provide new insights in digital pathology.

In this work, we introduce \ours, a novel and efficient attention-based MIL framework for WSI analysis, proposing a paradigm shift by removing the complexity of correlation learning from the MIL aggregator, using context-aware patch representations. 
Following the architecture of Transolver \cite{wu_transolver_2024, luo_transolver_2025}, which shows promising results in efficient PDE modeling, we leverage Multi-Head Self-Attention (MSA) over a small set of global context-aware tokens, achieving linear computational complexity with respect to the input and promoting effective correlation learning on downstream tasks. More precisely, our main contributions are summarized as follows:\newline
\noindent\textbf{(i) We propose a novel and efficient MIL setting based on the Transolver architecture.} Tackling the challenge of the large dimensionality of the input, \ours introduces a bottleneck before the attention operator, which consists of: (1) soft clustering of the patch embeddings and (2) aggregating each cluster into a context-aware token. By utilizing MSA over the context-aware tokens, \ours achieves linear computational complexity with respect to the bag size and produces rich morphology/context-aware patch representations
.\newline
\textbf{(ii) A highly parameter-efficient formulation.} Our approach performs on par with current state-of-the-art MIL heads, while reducing the total number of trainable parameters by 48\% compared to ABMIL and up to 92.8\% compared to SOTA transformer-based MILs. This significantly reduces the computational requirements during training and inference in terms of time, FLOPS, and memory utilization.\newline
\textbf{(iii) A scalable, aggregator-agnostic formulation that can be adapted in multiple MIL heads}. Our formulation is independent of the MIL aggregator, and it can be applied in different commonly used MIL settings with small computational overhead.

We challenge \ours on various publicly available computational pathology datasets. Paired with a simple MeanMIL aggregator, our method matches SOTA performance, while achieving leading efficiency, highlighting a highly efficient and adaptable MIL framework. 

\vspace{-2mm}
\section{Related Work}

\paragraph{MIL-based frameworks for digital pathology.} 
During the last years, many different MIL settings have been introduced and extensively tested in different settings~\cite{miltransfer_mahmood_icml}. Depending on the mechanism of aggregation that they are using, they can be grouped into different categories. Among the most popular attention-based methods, we can note ABMIL~\cite{ilse_attention-based_2018}, CLAM~\cite{lu_data-efficient_2021} and DSMIL~\cite{li_dual-stream_2021}. Moreover, TransMIL~\cite{shao_transmil_2021} was among the first to introduce a transformer network specifically for WSI, in order to model both morphological and spatial correlations. Building on top of this, DGRMIL \cite{zhu_dgr-mil_2025} utilizes a set of learnable ”global vectors” to summarize distinct morphological patterns and computes cross-attention between the instances and these global vectors, effectively achieving linear scaling.
Finally, probabilistic-based MIL such as the one introduced at Cui et al.~\cite{cui_bayes-mil_2022} argued that standard attention scores are unreliable proxies for interpretability, proposing BayesMIL, where they introduce a probabilistic instance-wise attention module that yields patch-level uncertainty estimates. Similarly,   Lolos et al. targeted the lack of uncertainty estimation in deterministic models and introduced SGPMIL~\cite{lolos_sgpmil_2025}, a framework to learn a posterior distribution over attention scores.

\paragraph{Attention-based Neural PDE Solvers.}
Solving Partial Differential Equations (PDEs) is fundamental to modeling complex phenomena in science and engineering.
While traditional numerical approaches such as the Finite Element Method (FEM) offer high accuracy, they typically require discretization of the domain into high-resolution meshes -often containing millions of mesh points-, resulting in prohibitive computational costs
~\cite{grossmann_can_2024}. Consequently, deep learning-based neural operators have emerged as efficient surrogates, capable of learning the mapping between model state and solution fields directly from data \cite{li_fourier_2020, lu_learning_2021, wu_transolver_2024}.
Transformer architectures have been increasingly utilized in neural PDE solvers due to their ability to model global dependencies \cite{li_transformer_2022}. However, they often face computational bottlenecks due to the quadratic complexity of standard self-attention \cite{katharopoulos_transformers_2020, luo_transolver_2025}. Furthermore, simply applying attention to individual mesh points may fail to capture the intricate high-order physical correlations governing the system, as the model can become overwhelmed by low-level geometric details, 
thus preventing effective relation learning \cite{wu_flowformer_2022}. 
We identify that challenges inherent to long-sequence processing, such as computational complexity and efficient correlation learning, are common in both large-scale physical simulations and WSI analysis. Surprisingly, the use of neural PDE solvers has not been explored in digital pathology, to the best of our knowledge.

\paragraph{The Transolver Architecture.}
To address the prohibitive computational cost and degraded correlation learning due to the large size of the input, Wu et al. introduced Transolver \cite{wu_transolver_2024}, a "Transformer-based PDE solver for General Geometries", which was later scaled by Luo et al.~\cite{luo_transolver_2025}. Their architecture introduces Physics-Attention, proposing that a domain discretized to $N$ mesh points can be decomposed into a set of  $M \ll N$ physically consistent clusters (“slices”), which can then be aggregated into “physics-aware tokens”, forming a compact latent representation of distinct physical states. Standard Multi-Head Self-Attention (MSA) can then be applied to these tokens for correlation modeling 
with complexity $O(M^2)$, achieving linear scaling with respect to the number of mesh points. By explicitly modeling "physical states" rather than individual points, the model becomes more robust to geometric variations and discretization artifacts \cite{wu_transolver_2024, luo_transolver_2025}, while the learned slices have been shown to correspond to meaningful physical regions, enhancing the model's interpretability and generalization capability \cite{wu_transolver_2024, luo_transolver_2025}. 
Drawing a parallel to digital pathology, both neural PDE solvers and MIL models face the fundamental challenge of efficiently learning correlations over massive sequences of instances (mesh points in PDEs, patches in WSIs). Viewed through this lens, Transolver's Physics-Attention constitutes a promising approach to facilitate efficient global corelation modeling, by projecting the high-dimensional input space onto a compact set of latent variables.

\vspace{-2mm}
\section{Methodology}
In this work, we propose \ours, a novel and efficient MIL framework, designed to overcome the limitations of standard attention in WSI analysis. Unlike prior methods that treat patches as isolated units, \ours projects patch embeddings into morphology units via soft clustering, and aggregates them into a compact set of context-aware, low-dimensional global tokens, over which self-attention is performed. Global contextual information is then propagated back to the patch embeddings via context broadcasting.
By attending to the tokens rather than patch embeddings, \ours achieves linear scaling with respect to the bag size, while maintaining strong representational capacity and high parameter efficiency.

\subsection{Model Architecture}

\begin{figure}[ht]
\centering
\includegraphics[width=\textwidth]{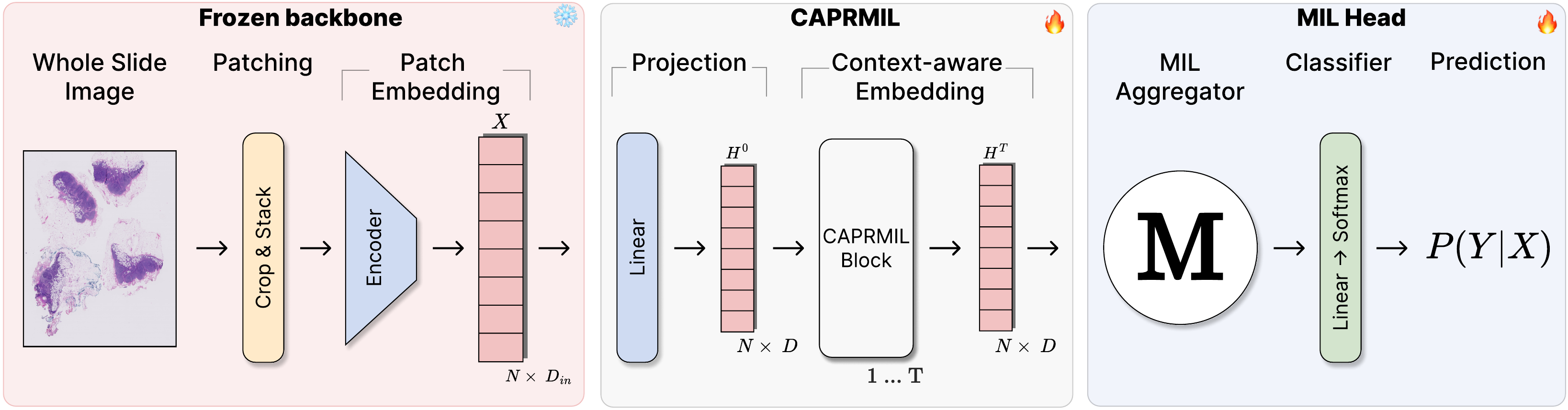}
\caption{\ours Framework. The WSI is tessellated into patches which are encoded into patch embeddings using a frozen backbone. After a linear projection, $T$ consecutive \ours Blocks augment global context to yield context-aware patch embeddings. A MIL aggregator and classifier then produce the final slide-level prediction.
}
\label{fig:\ours_framework}
\end{figure}

The \ours framework for WSI consists of three sequential stages (Figure~\ref{fig:\ours_framework}): (1) an initial projection of WSI patches into patch embeddings using a pre-trained encoder as frozen backbone, (2) a stack of \ours Blocks that use multi-head self-attention over global token representations to produce context/morphology-aware patch embeddings, and (3) a final MIL aggregation and classification head to produce the slide-level prediction. 

\subsubsection{Feature Projection}
A WSI is represented as a bag $X \in \mathbb{R}^{B \times N \times D_{in}}$ of $N$ patch embeddings of dimension $D_{in}$, with batch size $B$. These embeddings are projected into a latent space of dimension $D \ll D_{in}$ via a learnable linear layer followed by Layer Normalization (LN), GELU activation, and Dropout, yielding patch representations $\mathbf{H}^{(0)}$ as input to the first \ours Block:

$$
\mathbf{H}^{(0)}
= \text{Dropout}(\text{GELU}(\text{LN}(\text{Linear}(\mathbf{X})))) \in \mathbb{R}^{B \times N \times D}
$$

\subsubsection{The \ours Block}
\begin{figure}[t]
  \centering
  \begin{minipage}[t]{0.19\textwidth}
    \centering
    \includegraphics[width=\textwidth]{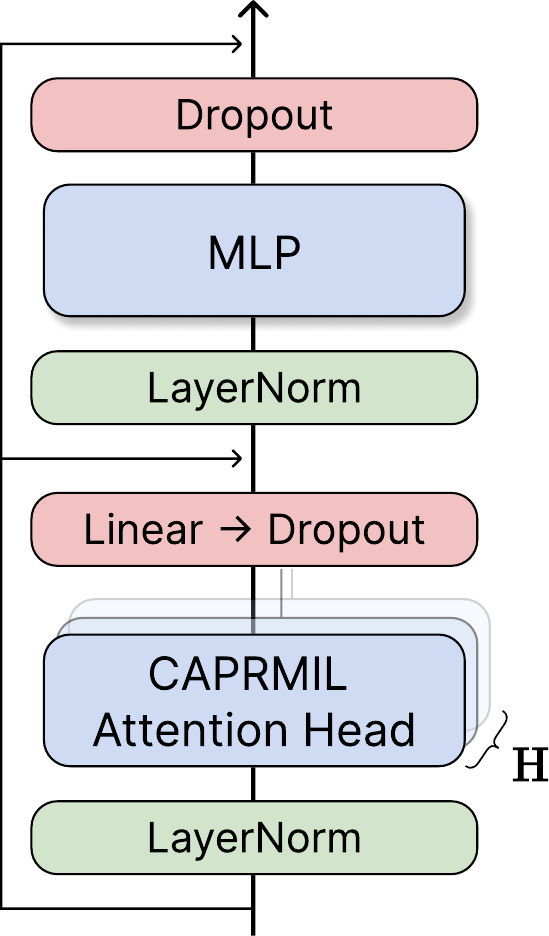}
    {\small (a) \ours Block}
  \end{minipage}
  \hfill
  \begin{minipage}[t]{0.78\textwidth}
    \centering
    \includegraphics[width=\textwidth]{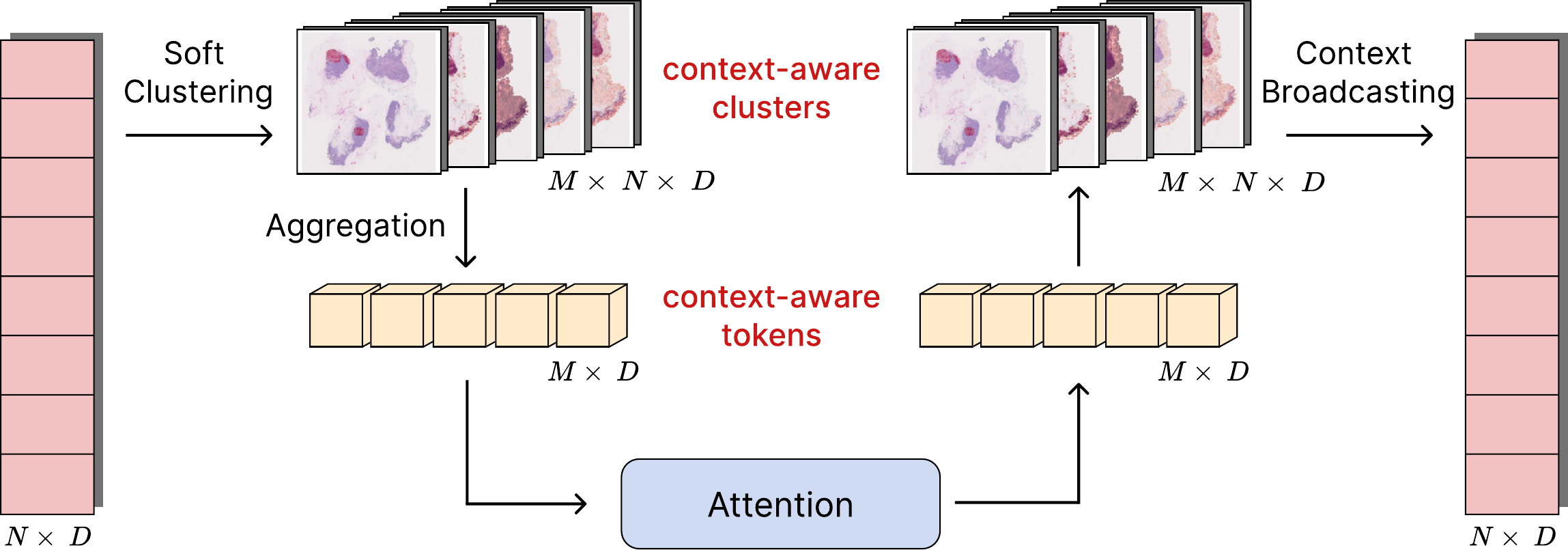}
    {\small (b) \ours Attention head}
  \end{minipage}
  \caption{(a) The \ours Block follows a transformer encoder architecture with multihead self-attention. Each \ours Block contains $H$ \ours Attention heads and their output is concatenated. Skip connections are implemented after every Dropout. (b) The \ours Attention head projects patch embeddings into $M$ clusters, via soft clustering. Each cluster is aggregated into a context-aware token and attention is applied to the set of $M$ tokens. The transited tokens are projected back to the input latent space via context broadcasting.
  }
  \label{fig:\ours_block_and_attention}
\end{figure}

To capture high-order correlations without the quadratic cost of standard self-attention,
the \ours\ Block adopts the Transolver architecture, performing attention over
low-dimensional, context-aware global tokens to achieve linear complexity with respect
to the bag size. As illustrated in Figure~\ref{fig:\ours_block_and_attention}a, it follows a
Transformer encoder-style design with $H$ \ours\ heads and shared projection matrices across heads, augmenting patch
embeddings with global context to produce rich, morphology-aware representations, formulated as:
$$
\mathbf{H}' = \mathbf{H}^{(l-1)} + \text{Dropout}(\text{\ours}\text{Attention}(\text{LN}(\mathbf{H}^{(l-1)})))
$$
$$
\mathbf{H}^{(l)} = \mathbf{H}' + \text{Dropout}(\text{MLP}(\text{LN}(\mathbf{H}')))
$$
for $l \in [1, T]$, for $T$ consecutive \ours Blocks. The MLP comprises two linear layers with GELU activation. \ours attention returns the concatenated output of all heads.

\subsubsection{\ours Attention head}
\ours adopts the Physics-Attention mechanism from Transolver to enable efficient correlation
learning on large-scale inputs. As illustrated in Figure~\ref{fig:\ours_block_and_attention}b, it operates in four stages: (1) soft clustering of patch
representations into morphology-aware clusters, (2) aggregation into morphology-aware tokens,
(3) self-attention over these tokens, and (4) broadcasting the transited tokens back to the
input space, producing context-aware patch representations.

\noindent\textbf{Soft Clustering.}
To achieve linear scaling with respect to the bag size, \ours replaces attention over patches
with attention over a compact set of morphology-aware tokens, produced via soft clustering
followed by aggregation.
Let $\mathbf{H} \in \mathbb{R}^{B \times N \times D}$ denote the input patch representations.
$\mathbf{H}$ is mapped via two learnable projections into 
$
\mathbf{x},\, \mathbf{f} \in
\mathbb{R}^{B \times N \times (H D_{\text{head}})},
\quad D_{\text{head}} = D / H,
$
by linear layers
$\mathbf{W}_x, \mathbf{W}_f \in \mathbb{R}^{D \times (H D_{\text{head}})}$
and reshaped into $\mathbf{\tilde{x}},\mathbf{\tilde{f}}\in\mathbb{R}^{B\times H\times N\times D_{\text{head}}}$. 
All $N$ patch embeddings are then softly assigned to $M$ context-aware clusters
per head. A learnable projection
$\mathbf{W}_{\text{cluster}} \in \mathbb{R}^{D_{\text{head}} \times M}$,
initialized orthogonally, produces:
\[
\mathbf{A}_{\text{logits}} = \frac{\mathbf{\tilde{x}} \mathbf{W}_{\text{cluster}}}{\tau},
\quad
\mathbf{W} = \text{Softmax}(\mathbf{A}_{\text{logits}}, \text{dim}=-1),
\quad
\sum_{m=1}^{M} W_{b,h,n,m} = 1 \ \forall\, b,h,n,
\]
where $\mathbf{A}_{\text{logits}}, \mathbf{W} \in \mathbb{R}^{B \times H \times N \times M}$,
$\mathbf{W}$ is the soft assignment matrix, corresponding to the probability distribution of each embedding being mapped to each cluster. $M$ controls the number of clusters, and $\tau \in \mathbb{R}^H$ is a temperature parameter we introduce as learnable, controlling the assignment entropy of each attention head. Each cluster is then aggregated into a token, corresponding to a weighted combination of input embeddings, calculated as:
\[
\mathbf{S}_{b,h,m}
=
\frac{\sum_{n=1}^{N} W_{b,h,n,m}\, \mathbf{\tilde{f}}_{b,h,n}}
     {\sum_{n=1}^{N} W_{b,h,n,m} + \varepsilon},
\quad
\text{with  }\mathbf{S} \in \mathbb{R}^{B \times H \times M \times D_{\text{head}}}
\]

\noindent\textbf{Self-Attention}.
Given $M$ morphology-aware tokens $\mathbf{S}$ per head, we apply Multi-Head Self-Attention. Query ($\mathbf{Q}$), Key ($\mathbf{K}$), and Value ($\mathbf{V}$) are obtained via shared linear projections of the head-wise token embeddings $\mathbf{S}$. Attention is then given by:
\[
\text{Attn} = \text{Softmax}\!\left(\frac{\mathbf{Q}\mathbf{K}^\top}{\sqrt{D_{\text{head}}}}\right),
\quad \mathbf{S}' = \text{Dropout}(\text{Attn} \cdot \mathbf{V})
\]
with $\mathbf{S}'\in \mathbb{R}^{B \times H \times M \times D_{head}}$. Since  ($M \ll N$), applying the attention operator over the context-aware tokens -instead of the $N$ patch embeddings- reduces computational complexity and allows the model to scale linearly with the input. At the same time, since tokens aggregate global context, \ours learns meaningful correlations, beyond spatial features.

\noindent\textbf{Context Broadcasting}.
The updated tokens $\mathbf{S}'$ are broadcast back to the input latent space using the same
assignment weights $\mathbf{W}$ from the soft clustering step, reconstructing each patch
representation as a weighted combination of transited tokens:
\[
\mathbf{O}_{b,h,n,d}
=
\sum_{m=1}^{M} \mathbf{S}'_{b,h,m,d}\, \mathbf{W}_{b,h,n,m},
\quad
\mathbf{O} \in \mathbb{R}^{B \times H \times N \times D_{\text{head}}}.
\]
Head-wise representations are concatenated into $\mathbf{H}^{(T)}\in\mathbb{R}^{B\times N\times (HD_{head})}$ and linearly projected to the
model dimension, yielding the final context-aware patch representations.

\subsubsection{Aggregation and Prediction}
After $T$ \ours Blocks, the patch representations $\mathbf{H}^{(T)}$ are mean-pooled to form a slide-level embedding $\mathbf{z} = \frac{1}{N} \sum_{n=1}^{N} \mathbf{H}^{(T)}_n \in \mathbb{R}^{B \times D}$. 
A final linear classifier projects $\mathbf{z}$ to the target class logits depending on the task.

\subsection{Computational Efficiency}
\ours addresses the "curse of dimensionality" by decoupling the sequence length $N$ from the attention mechanism. 
Since the attention operator displays quadratic computational complexity, attending to all $N$ patch embeddings would yield $O(N^2)$ complexity. \ours Attention instead attends to the $M$ context-aware tokens, achieving an overall complexity of $O(MND + M^2D)$. Given that the number of prototypes $M$ is a constant with $M\ll N$, the model achieves linear computational complexity with respect to the input size $N$, making it ideal to model long sequences.


\vspace{-3mm}
\section{Experiments \& Results}
\paragraph{Datasets, Tasks and Evaluation Metrics.}
We evaluate our approach on four WSI benchmarks: \textbf{CAMELYON16}~\cite{ehteshami_bejnordi_diagnostic_2017}
for tumor detection, \textbf{TCGA-NSCLC}~\cite{cooper_pancancer_2018,campbell_distinct_2016}
for lung cancer subtyping, \textbf{BRACS}~\cite{brancati_bracs_2022}
for coarse breast lesion classification, and \textbf{PANDA}~\cite{panda_dataset}
for prostate ISUP grading. Dataset-specific evaluation protocols, metrics, and implementation
details are provided in the Appendix. We report slide-level classification performance and
calibration using area under the curve (\textbf{AUC}) and adaptive expected calibration error
(\textbf{ACE})~\cite{nixon_measuring_2019}.

\subsection{Slide-level Performance and Parameter Efficiency}

\begin{table*}[htbp]
\centering
\small
\setlength{\tabcolsep}{6pt}
\resizebox{\textwidth}{!}{
\begin{tabular}{lcccccccccc}
\toprule
 & \multicolumn{2}{c}{CAMELYON16} 
 & \multicolumn{2}{c}{TCGA-NSCLC} 
 & \multicolumn{2}{c}{PANDA} 
 & \multicolumn{2}{c}{BRACS} 
 & Params & FLOPs \\
\cmidrule(lr){2-3}
\cmidrule(lr){4-5}
\cmidrule(lr){6-7}
\cmidrule(lr){8-9}

& AUC & ACE 
& AUC & ACE 
& $\kappa$ & ACE 
& AUC & ACE 
& (M) & (G) \\
\midrule

ABMIL~\cite{ilse_attention-based_2018}   
& $\mathbf{.987_{.005}}$ & $.036_{.004}$
& $.973_{.009}$ & $.039_{.008}$
& $.910_{.028}$ & $.044_{.015}$
& $\underline{.852_{.025}}$ & $\underline{.175_{.007}}$
& $.660$ & $1.31$ \\

CLAM~\cite{lu_data-efficient_2021}      
& $\underline{.986_{.004}}$ & $.044_{.027}$
& $.953_{.004}$ & $.056_{.016}$
& $.927_{.025}$ & $.031_{.018}$
& $.850_{.021}$ & $.183_{.011}$
& $.920$ & $1.84$ \\

TransMIL~\cite{shao_transmil_2021}
& $.978_{.004}$ & $.044_{.012}$
& $.970_{.012}$ & $.046_{.019}$
& $.911_{.030}$ & $.043_{.021}$
& $.826_{.032}$ & $.186_{.012}$
& $2.67$ & $85.02$ \\

DGRMIL~\cite{zhu_dgr-mil_2025}
& $.967_{.018}$ & $.027_{.021}$
& $.974_{.011}$ & $\underline{.038_{.022}}$
& $.933_{.047}$ & $.036_{.025}$
& $.818_{.035}$ & $.186_{.023}$
& $4.34$ & $79.88$ \\

BayesMIL~\cite{cui_bayes-mil_2023}
& $.975_{.006}$ & $\mathbf{.023_{.006}}$
& $.973_{.021}$ & $\mathbf{.033_{.017}}$
& $.926_{.031}$ & $.031_{.016}$
& $.829_{.022}$ & $.183_{.028}$
& $1.32$ & $2.63$ \\



SGPMIL~\cite{lolos_sgpmil_2025}    
& $\mathbf{.987_{.008}}$ & $\underline{.026_{.009}}$
& $.973_{.014}$ & $.047_{.027}$
& $\mathbf{.955_{.037}}$ & $\underline{.028_{.022}}$
& $\mathbf{.870_{.026}}$ & $\mathbf{.142_{.032}}$
& $1.21$ & $2.44$ \\
\hline

Mean
& $.693_{.046}$ & $.241_{.022}$ 
& $\mathbf{.979_{.015}}$ & $.041_{.019}$
& $.924_{.028}$ & $.035_{.013}$
& $.738_{.006}$ & $.223_{.015}$
& $\mathbf{.130}$ & $\mathbf{.260}$ \\

\textbf{\ours+Mean}     
& $.975_{.006}$ & $.028_{.006}$
& $\underline{.978_{.016}}$ & $\mathbf{.033_{.021}}$
& $\underline{.944_{.053}}$ & $\mathbf{.021_{.024}}$
& $.850_{.031}$ & $.189_{.026}$
& $\underline{.314}$ & $\underline{.628}$ \\

\bottomrule
\end{tabular}}
\caption{Slide-level performance comparison across datasets. Results are reported as AUC/$\kappa$, and ACE. FLOPs are measured per forward pass for a bag of 1000 patch embeddings at inference. Using a mean operator in the initial projection layer before classification (i.e., our approach without the block) leads to substantial performance degradation for large bag sizes, such as CAMELYON16 and BRACS.}
\label{tab:bag-level-performance_uni}
\end{table*}

\ours achieves competitive AUC and calibration relative to state-of-the-art MIL methods
(Table~\ref{tab:bag-level-performance_uni}), with performance differences consistently
within one standard deviation, while being substantially more parameter efficient.
Notably, these results are obtained using simple mean aggregation, underlining the
strength of the learned context-aware patch representations. Across datasets, \ours
matches parameter-efficient methods such as ABMIL and CLAM on CAMELYON16 and
TCGA-NSCLC, and ranks among the top-performing approaches on multiclass tasks including
PANDA and BRACS. At the same time, \ours reduces trainable parameters by approximately
48\% relative to ABMIL, and by up to 88\% and 92.8\% compared to transformer-based methods
such as TransMIL and DGRMIL, respectively. In terms of efficiency, \ours reduces FLOPS during inference by 52\% to over 99\% compared to ABMIL, TransMIL, and DGRMIL.

In contrast, naive mean aggregation -using a linear projection, mean pooling, and a
linear classifier- degrades substantially on large-bag tasks. The Mean
baseline underperforms by 44\% on CAMELYON16 and 35.5\% on BRACS, where bags contain
4k–20k instances. On PANDA (average bag size $\sim$500), mean pooling performs comparably
to other methods, with a similar trend on TCGA-NSCLC. Overall, these results indicate that context-aware tokenization is critical for
maintaining discriminative capacity while enabling a parameter- and computation-efficient
formulation.

\begin{figure}[t]
    \centering
    \includegraphics[width=1\linewidth]{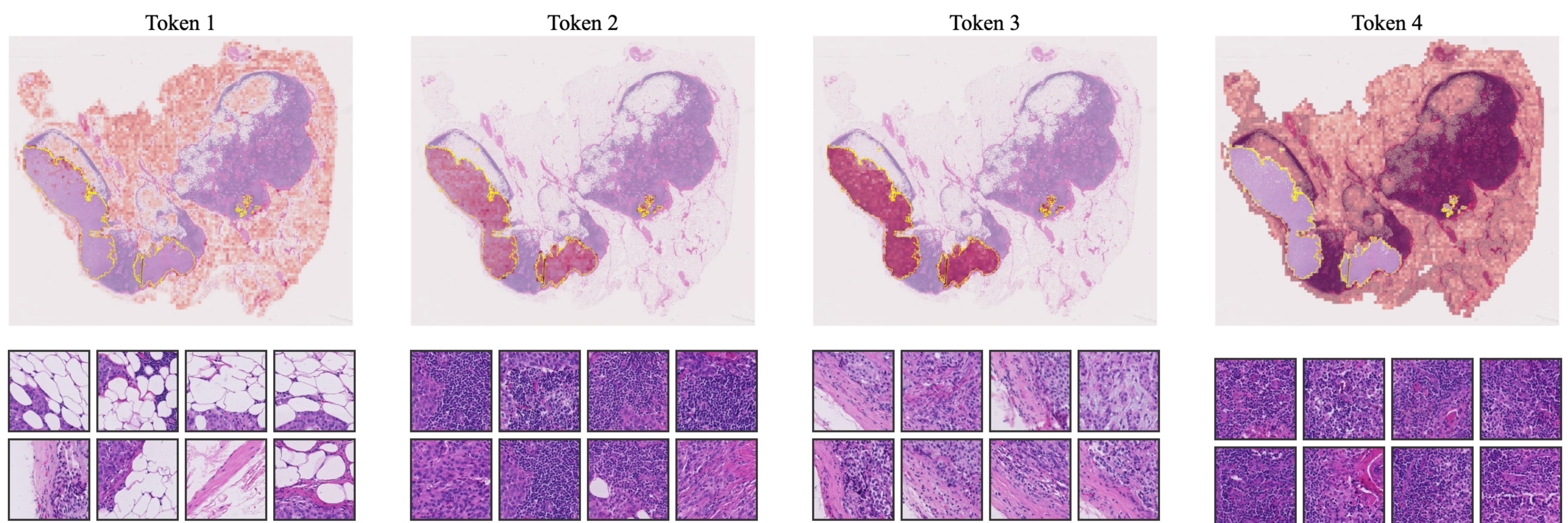}
    \caption{\textbf{Token–patch assignment heatmaps.} Test slide from CAMELYON16. \textit{Top:} Soft assignment weights from one \ours attention head, indicating each patch’s contribution to the $M$ context-aware tokens. \textit{Bottom:} Top-$8$ patches per token ranked by assignment score, highlighting dominant morphological patterns for each token.}
    \label{fig:test_016_heatmap}
\end{figure}

As seen in Figures~\ref{fig:test_016_heatmap} and A~\ref{fig:test_001_heatmap} -~\ref{fig:test_075_heatmap}, we observe that tokens tend to aggregate patches with visually coherent histological patterns, such as adipose-rich or epithelial-dominant regions, while de-emphasizing unrelated tissue types. 
The top-$k$ assigned patches per token indicate that a limited subset of instances dominates each token’s 
construction. 
Specifically in Figure~\ref{fig:test_016_heatmap}, Token~1 predominantly captures adipose-rich regions, as confirmed by their
low cellular content in the top-8 assigned patches. Tokens~2 and~3 focus on tumor-related
tissue, with Token~2 aggregating malignant epithelial regions, while Token~3 captures
stromal or tumor-associated connective tissue. Finally, Token~4 primarily represents benign
tissue, with patches exhibiting more homogeneous cellular organization.

\subsection{Computational and Memory Efficiency}

\begin{figure}[h]
    \centering
    \subfigure[
    ]{
        \includegraphics[width=0.46\linewidth]{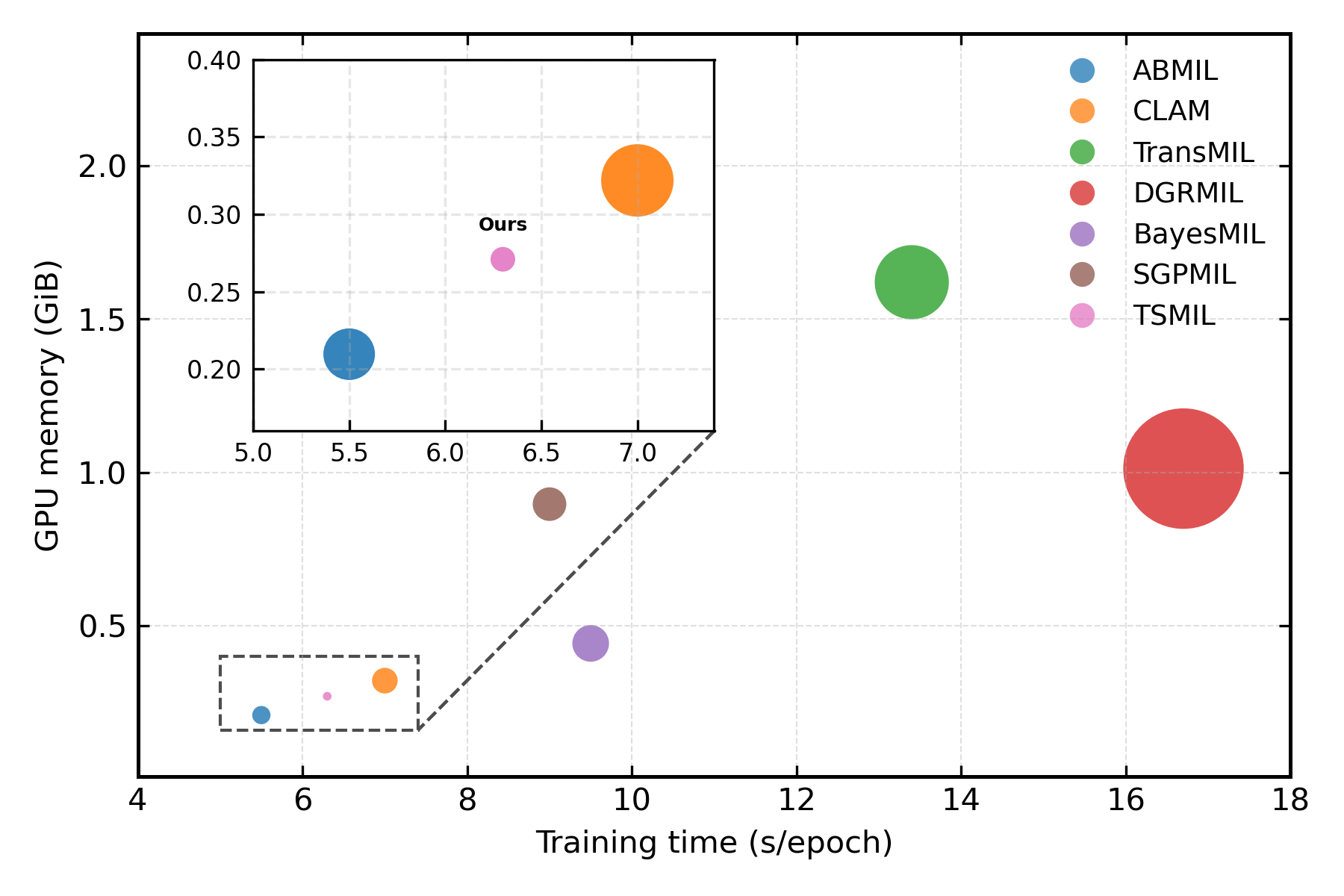}
        \label{fig:model_efficiency_scatterplot}
    }
    \hfill
    \subfigure[
    ]{
        \includegraphics[width=0.46\linewidth]{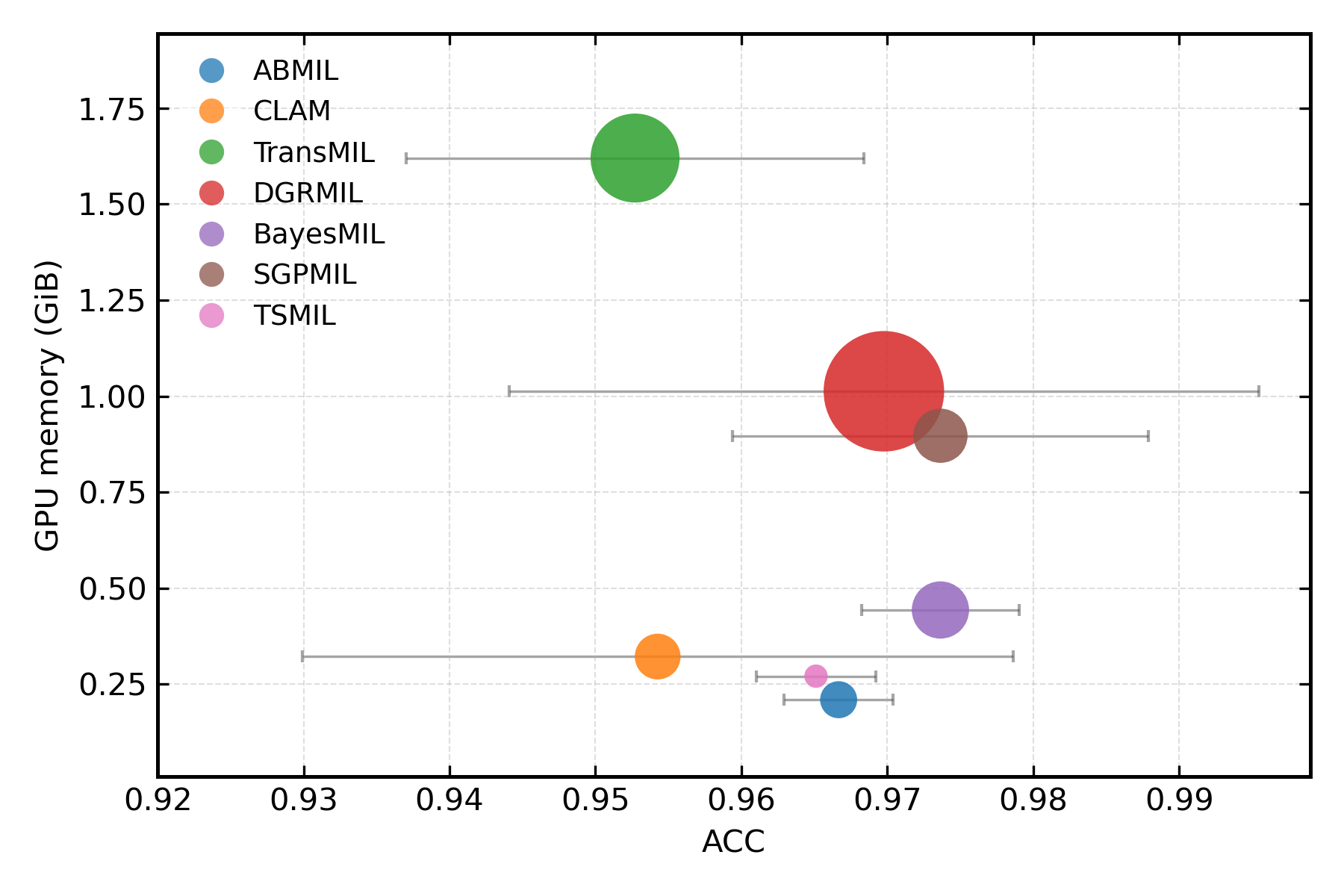}       \label{fig:model_efficiency_vs_performance_scatterplot}
    }
    \caption{\textbf{Model efficiency analysis}.
    ($a$) GPU memory footprint (peak during training, averaged over 30 epochs) vs.\ training time
    (entire training set, averaged over 30 epochs).
    ($b$) GPU memory footprint vs.\ ACC.
    Marker size denotes the number of trainable parameters.}
    \label{fig:model_efficiency_combined}
\end{figure}

While parameter count and FLOPs provide useful proxies for model efficiency, practical deployment at whole-slide scale additionally depends on empirical resource utilization. Figure~\ref{fig:model_efficiency_combined} analyzes this by relating peak GPU memory usage, training time, and slide-level performance across competing MIL methods. As shown in Figure~\ref{fig:model_efficiency_combined}(a), \ours exhibits a substantially lower memory footprint and shorter training time compared to transformer-based approaches, reflecting its linear-scaling design and low-dimensional intermediate representations. \ours remains competitive with more computationally demanding models despite its resource-efficiency, by operating on rich, context-aware patch embeddings.
Figure~\ref{fig:model_efficiency_combined}(b) further illustrates the trade-off between accuracy and memory consumption. \ours achieves high balanced accuracy, with performance differences consistently within one standard deviation of leading competitors
, while operating under a significantly smaller GPU memory budget. In contrast, transformer-based methods such as TransMIL and DGRMIL incur large memory overheads for only marginal performance gains. While attention-based and probabilistic MIL methods offer stronger aggregation modules, they do so with increased computational or memory requirements, suggesting that \ours context-aware representations provide a lightweight yet competitive alternative.

\subsection{Ablation studies}

\paragraph{Clusters and heads.}
Varying the number of clusters $M$ while fixing $H=8$ and the MLP ratio to 4 shows stable performance across a wide range of values ($M\in\{2,4,8,16\}$), with no consistent gains from increasing the number of clusters beyond small to moderate values (Table A~\ref{tab:tsmil_ablations_main}, top). Similarly, increasing the number of attention heads $H$ improves performance from 2 to 8 heads but saturates thereafter, with no clear benefit on larger values (Table A~\ref{tab:tsmil_ablations_main}, middle). Based on these trends, we adopt $M=4$ and $H=8$ as balanced choices that provide sufficient contextual capacity without unnecessary complexity.

\paragraph{MLP expansion ratio.}
Ablating the MLP expansion ratio with fixed $M=4$ and $H=8$ indicates that smaller ratio slightly degrades performance, while larger ratio yields more consistent results across metrics (Table A~\ref{tab:tsmil_ablations_main}, bottom). We therefore use an MLP ratio of 4 in all experiments. Overall, these ablations indicate that the selected configuration ($M=4$, $H=8$, MLP ratio $=4$) offers a robust trade-off between representational capacity and efficiency and is not sensitive to precise hyperparameter choices.

\begin{table*}[htbp]
\centering
\small
\setlength{\tabcolsep}{6pt}
\resizebox{\textwidth}{!}{
\begin{tabular}{lccccccccccc}
\toprule
 & \multicolumn{2}{c}{CAMELYON16} 
 & \multicolumn{2}{c}{TCGA-NSCLC} 
 & \multicolumn{2}{c}{PANDA} 
 & \multicolumn{2}{c}{BRACS} 
 & Params \\
\cmidrule(lr){2-3}
\cmidrule(lr){4-5}
\cmidrule(lr){6-7}
\cmidrule(lr){8-9}

& AUC & ACE
& AUC & ACE
& $\kappa$ & ACE
& AUC & ACE
& (M) \\
\midrule

\textbf{\ours+Mean}     
& $.975_{.006}$ & $\underline{.028_{.006}}$
& $\mathbf{.978_{.016}}$ & $.033_{.021}$
& $.944_{.053}$ & $.021_{.024}$
& $\underline{.850_{.031}}$ & $.189_{.026}$
& $\mathbf{.314}$ \\

\textbf{\ours+Attn}     
& $\mathbf{.977_{.004}}$ & $\mathbf{.027_{.003}}$
& $\underline{.975_{.015}}$ & $\mathbf{.031_{.023}}$
& $\underline{.944_{.046}}$ & $.023_{.025}$
& $.834_{.032}$ & $\underline{.180_{.023}}$
& $\underline{.331}$ \\

\textbf{\ours+GAttn}     
& $\underline{.976_{.009}}$ & $.033_{.010}$
& $.974_{.018}$ & $\underline{.032_{.020}}$
& $\mathbf{.952_{.043}}$ & $\mathbf{.019_{.022}}$
& $\mathbf{.874_{.031}}$ & $\mathbf{.171_{.019}}$
& $.347$ \\

\bottomrule
\end{tabular}

}
\caption{
Comparison of aggregation strategies within \ours.
Results are reported as AUC/$\kappa$, and ACE; parameter counts include the aggregation module.
}
\label{tab:modularity_uni}
\end{table*}

\vspace{-2mm}
\paragraph{Modularity and aggregation robustness.} Table~\ref{tab:modularity_uni} compares different MIL aggregation strategies learned together with \ours representations.
In contrast to prior MIL approaches that rely heavily on sophisticated attention pooling, we observe that replacing mean aggregation with attention or gated attention leads to broadly comparable performance across datasets, within one standard deviation. Notably, on more challenging multiclass tasks such as PANDA and BRACS, attention-based aggregators yield a performance increase from $0.8\%$ up to $2.4\%$ respectively, suggesting that additional aggregation capacity may be beneficial in more complex settings.
Overall, these results indicate that the \ours block already encodes most of the relevant contextual and discriminative information at the patch level, rendering the choice of final aggregation largely non-critical for performance.
While attention-based aggregators introduce increased parameterization, they do not provide consistent gains across all tasks, highlighting diminishing returns once strong instance representations are learned.
These findings underline the modularity of \ours and demonstrate that competitive performance can be achieved with simple, parameter-efficient aggregation, while still allowing the flexibility to incorporate more expressive MIL heads when task complexity demands it.

\vspace{-3mm}
\section{Conclusions}
We present a parameter-efficient and scalable MIL framework that learns context-aware patch representations, substantially reducing reliance on complex aggregation mechanisms. Experimental results show that once rich contextual features are learned, simple pooling performs on par with more elaborate MIL heads, underscoring the robustness and modularity of the proposed approach. A current limitation is the focus on unimodal visual inputs; evaluating scalability and robustness in larger multimodal pipelines remains and interesting direction for future work.

\clearpage  
\midlacknowledgments{This work has been partially supported by project MIS 5154714 of the National Recovery
and Resilience Plan Greece 2.0 funded by the European Union under the NextGenerationEU Program.
Hardware resources were granted with the support of GRNET.}

\bibliography{midl-samplebibliography}

\appendix
\newpage

\section{Experiments}
\subsection{Datasets}
\label{sec:datasets}
\paragraph{CAMELYON16}~\cite{ehteshami_bejnordi_diagnostic_2017} consists of 399 WSIs of sentinel lymph node tissue sections derived from women with breast cancer. The dataset is split into a training set of 270 images and a test set of 129 images. Collected from two medical centers in the Netherlands, it includes exhaustive pixel-level annotations of metastatic regions (both macrometastases and micrometastases) verified by expert pathologists. We use TRIDENT \cite{vaidya_molecular-driven_2025, zhang_accelerating_2025} to segment and patch the WSIs at 10× magnification~\cite{mammadov_self-supervision_2025} into 224x224 non-overlapping patches and utilize the UNIv1~\cite{chen_towards_2024} encoder for feature extraction \cite{chen_towards_2024} . Similarly to Lu et. al \cite{lu_data-efficient_2021}, we follow a 10-fold cross-validation protocol and report mean bag-level performance.

\paragraph{TCGA-NSCLC} We use the dataset from The Cancer Genome Atlas (TCGA) program for the non-small cell lung carcinoma (NSCLC) subtyping task~\cite{cooper_pancancer_2018, campbell_distinct_2016}. The dataset consists of Hematoxylin and Eosin (H\&E) stained WSIs in 2 distinct cohorts: Lung Adenocarcinoma (TCGA-LUAD) and Lung Squamous Cell Carcinoma (TCGA-LUSC) \cite{campbell_distinct_2016, cooper_pancancer_2018}. Specifically, we use 494 LUAD and 512 LUSC cases for a total of 1,006 slides, segment and patch at 10× magnification \cite{mammadov_self-supervision_2025} into 224x224 non-overlapping patches and use the UNIv1~\cite{chen_towards_2024} encoder for feature extraction. Performance is reported over 4 folds.

\paragraph{PANDA}~\cite{panda_dataset} is derived from the MICCAI 2020 Prostate Cancer Grade Assessment challenge and comprises 10,609 WSIs from prostate core needle biopsies annotated, providing slide-level Gleason scores and ISUP grades alongside expert tissue annotations. We address ISUP grading (0-5) as a 6-class classification task and follow a 5-fold cross-validation protocol using stratified splits, with each fold containing approximately 80 splits for training, 5 for validation and 15 for testing. We segment the WSIs into non-overlapping patches of size 224×224 pixels at 20× magnification \cite{song_morphological_2024} and use the UNIv1~\cite{chen_towards_2024} encoder for feature extraction.

\paragraph{BRACS}~\cite{brancati_bracs_2022} The BReAst Carcinoma Subtyping (BRACS) dataset comprises 547 H\&E stained WSIs and over 4,500 annotated regions of interest derived from 189 patients, designed to advance the automatic detection of challenging "atypical" (precancerous) lesions that are often underrepresented in other public datasets. It is annotated into seven histological subtypes, grouped into three main categories: Benign (Normal, Pathological Benign, Usual Ductal Hyperplasia), Atypical (Flat Epithelial Atypia, Atypical Ductal Hyperplasia), and Malignant (Ductal Carcinoma in Situ, Invasive Carcinoma). We specifically focus on coarse classification into the three main categories (3-class classification), using the train/validation/test split provided with the dataset. We segment the WSIs into non-overlapping patches of size 224×224 pixels at 20× magnification \cite{song_morphological_2024}, and use the UNIv1~\cite{chen_towards_2024} encoder for feature extraction. Performance is reported over 5 seeds.

\subsection{Implementation Details}
\label{sec:implemntation_details}
All models are trained and evaluated in Python with PyTorch, using the same PyTorch Lightning
training pipeline with identical data loading, batching, and hardware configurations.
For \ours, training is performed using the standard cross-entropy loss on slide-level labels,
while competing methods are trained using the loss functions specified in their original works.

The models are trained for a maximum of 30 epochs on a single A100 GPU, using full-precision (FP32) arithmetic, except MeanMIL which is trained for a maximum of 50 epochs to obtain convergence. \ours optimization employs AdamW with a base learning rate of $2\times10^{-4}$, weight decay of $1\times10^{-5}$, and momentum parameter $0.9$. We use a cosine annealing learning-rate schedule, with a 6-epoch warm-up phase starting at $1\times10^{-5}$, and minimum learning rate of $1\times10^{-7}$. Early stopping was governed by a patience of 20 epochs and a performance threshold of $10^{-4}$. Learning-rate dynamics were logged every epoch, with explicit tracking of weight-decay values to enable fine-grained monitoring of the training process.

Model-specific hyperparameters and optimizer choices for competitors follow the respective original papers and are selected to ensure stable convergence based on observed loss curves. FLOPs are reported for a single forward pass during evaluation and a dummy input bag of 1000 patch embeddings and serve as a proxy for algorithmic complexity. Wall-clock training and inference times measure end-to-end execution. We additionally report peak GPU utilization as an implementation-level efficiency metric reflecting how effectively each model translates computation into hardware usage under identical experimental conditions. Complete code and instructions are publicly available at \url{https://github.com/mandlos/CAPRMIL}.

\subsection{Evaluation of context-aware tokens}
\begin{figure}[ht]
    \centering
    \includegraphics[width=1\linewidth]{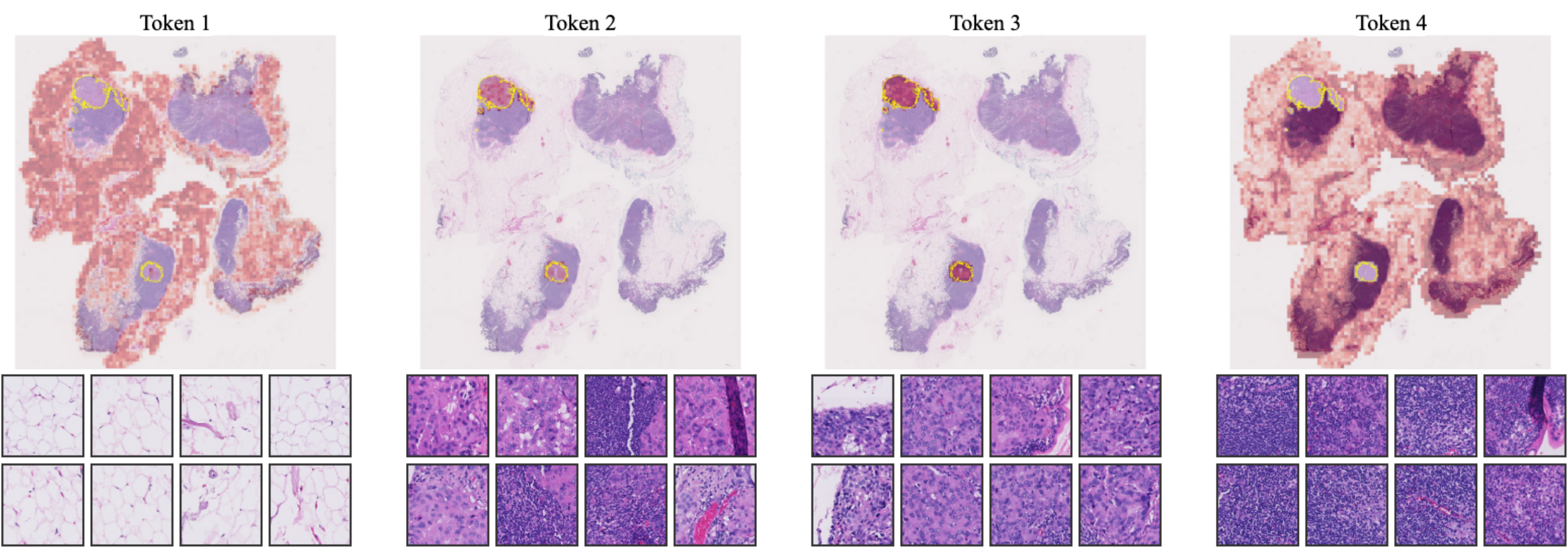}
    \caption{\textbf{Token–patch assignment heatmaps.} Test slide from CAMELYON16. \textit{Top:} Soft assignment weights from one TSMIL attention head, indicating each patch’s contribution to the $M$ context-aware tokens. \textit{Bottom:} Top-$8$ patches per token ranked by assignment score, highlighting the dominant morphological patterns contributing to each token.}
    \label{fig:test_001_heatmap}
\end{figure}

\begin{figure}[ht]
    \centering
    \includegraphics[width=1\linewidth]{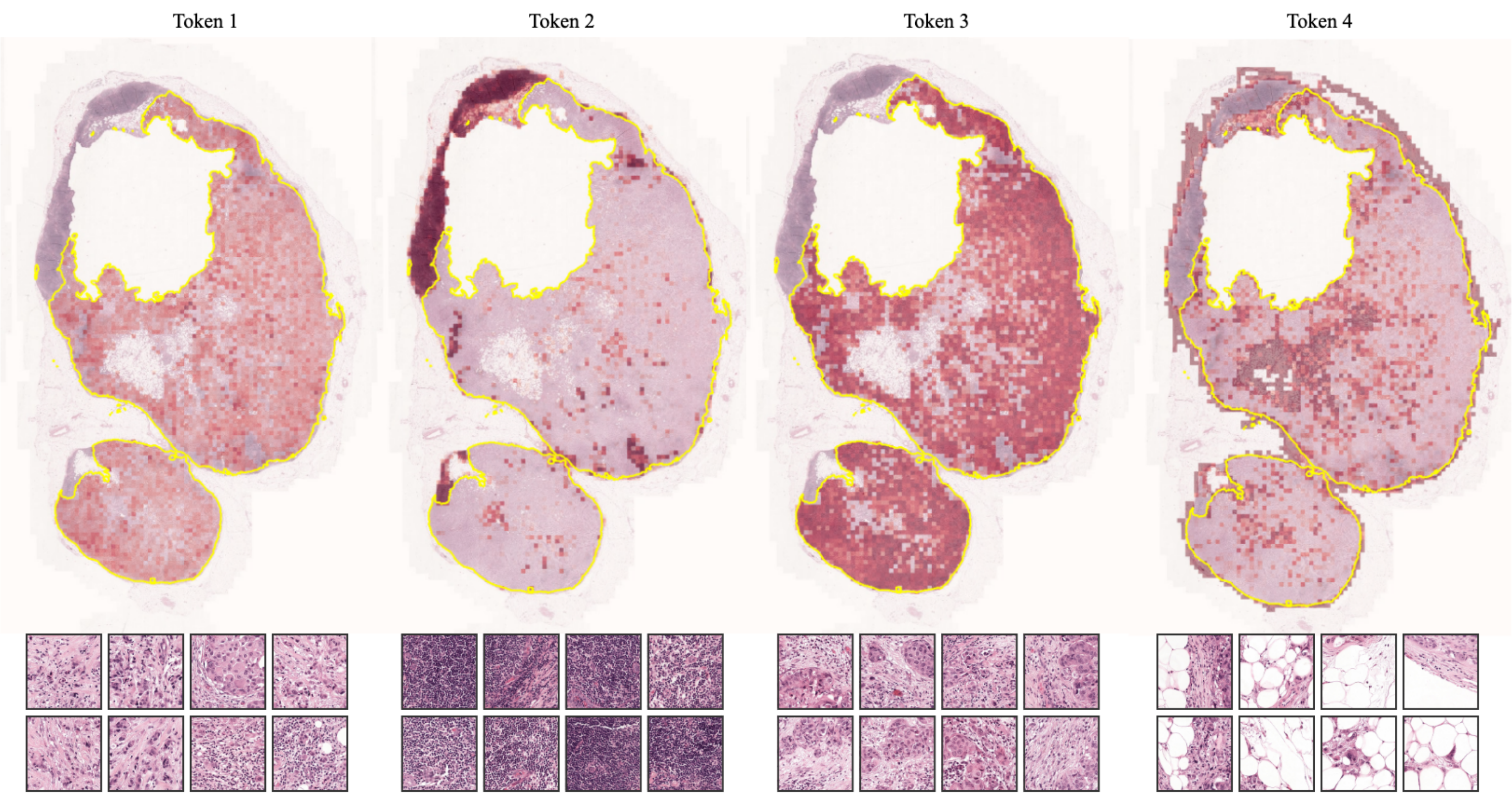}
    \caption{\textbf{Token–patch assignment heatmaps.} Test slide from CAMELYON16. \textit{Top:} Soft assignment weights from one TSMIL attention head, indicating each patch’s contribution to the $M$ context-aware tokens. \textit{Bottom:} Top-$8$ patches per token ranked by assignment score, highlighting the dominant morphological patterns contributing to each token.}
    \label{fig:test_021_heatmap}
\end{figure}

\begin{figure}[ht]
    \centering
    \includegraphics[width=1\linewidth]{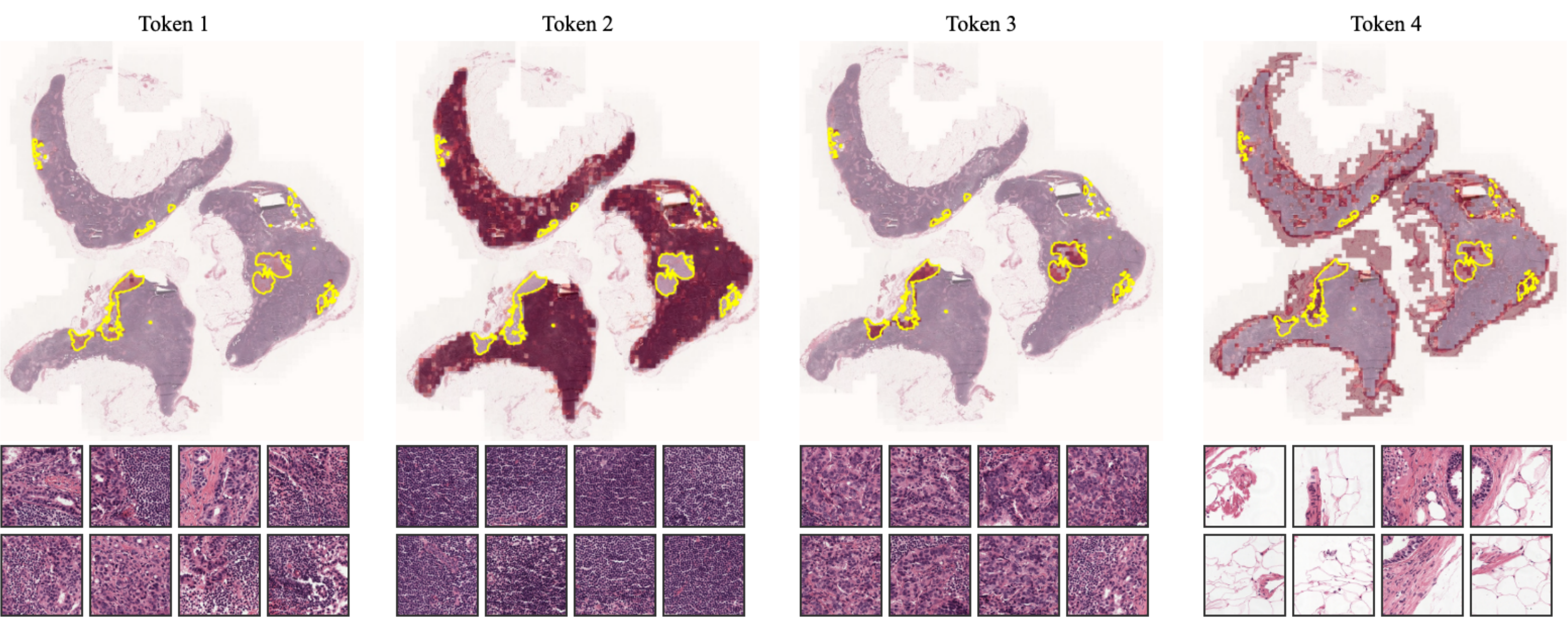}
    \caption{\textbf{Token–patch assignment heatmaps.} Test slide from CAMELYON16. \textit{Top:} Soft assignment weights from one TSMIL attention head, indicating each patch’s contribution to the $M$ context-aware tokens. \textit{Bottom:} Top-$8$ patches per token ranked by assignment score, highlighting the dominant morphological patterns contributing to each token.}
    \label{fig:test_075_heatmap}
\end{figure}

\section{Ablation Studies}
\begin{table}[h]
\centering
\small
\begin{tabular}{cccccc}
\toprule
Clusters (M) & Heads (H) & MLP ratio & AUC & ACE & Params (M) \\
\midrule
2  & 8  & 4 & $.971_{.009}$ & $\mathbf{.027_{.007}}$ & $0.314$\\
4  & 8  & 4 & $\mathbf{.975_{.006}}$ & $.028_{.006}$ & $0.314$ \\
8  & 8  & 4 & $.974_{.009}$ & $.030_{.008}$ & $0.314$ \\
16 & 8  & 4 & $.973_{.008}$ & $\mathbf{.027_{.008}}$ & $0.314$ \\
\midrule
4  & 2  & 4 & $.971_{.010}$ & $.031_{.011}$ & $0.326$ \\
4  & 4  & 4 & $.972_{.009}$ & $.033_{.011}$ & $0.316$ \\
4  & 8  & 4 & $\mathbf{.975_{.006}}$ & $\mathbf{.028_{.006}}$ & $0.314$ \\
4  & 12 & 4 & $.972_{.009}$ & $.032_{.006}$ & $\mathbf{0.310}$ \\
\midrule
4  & 8  & 1 & $.973_{.008}$ & $\mathbf{.027_{.006}}$ & $\mathbf{0.215}$ \\
4  & 8  & 2 & $.969_{.011}$ & $.030_{.008}$ & $0.248$ \\
4  & 8  & 4 & $\mathbf{.975_{.006}}$ & $.028_{.006}$ & $0.314$ \\
\bottomrule
\end{tabular}
\caption{
Ablation of the number of clusters $M$, attention heads $H$, and the MLP expansion ratio in the Transolver block.
Exhaustive ablation results for the number of clusters, attention heads, and input embedding dimensionality are provided in
Figures A~\ref{fig:ablation_clusters_cam16}, \ref{fig:ablation_heads_cam16_0}, \ref{fig:ablation_heads_cam16_1}, \ref{fig:ablation_hdims_cam16}, and \ref{fig:ablation_mlp_cam16}.
}
\label{tab:tsmil_ablations_main}
\end{table}

\begin{figure}[p]
\centering

\resizebox{0.625\textwidth}{!}{%
\begin{minipage}{\textwidth}

\includegraphics[width=\linewidth]{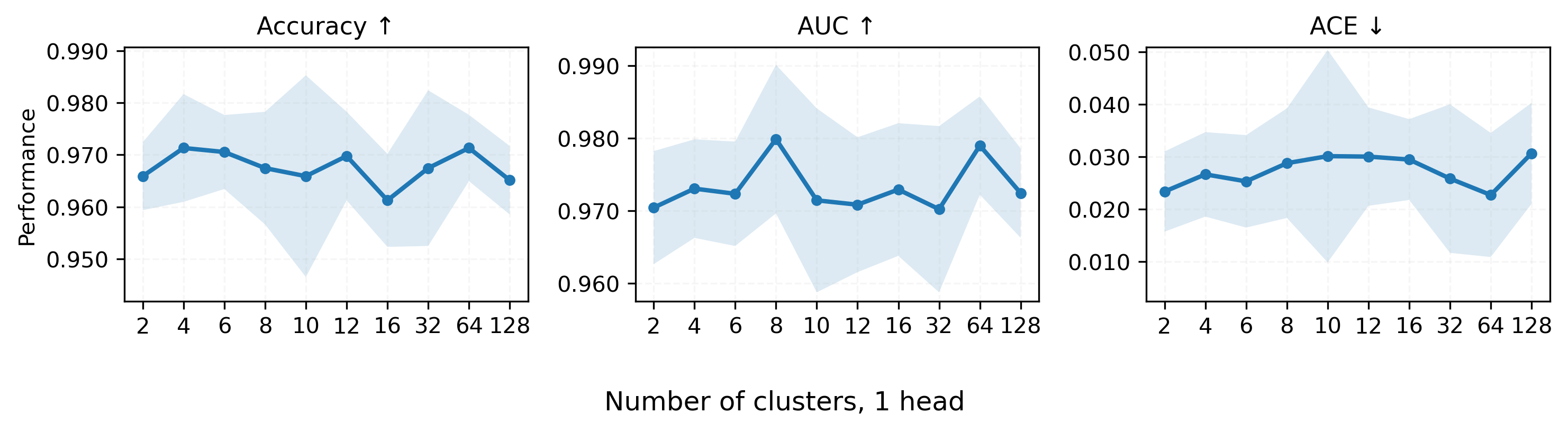}
\vspace{0.1em}

\includegraphics[width=\linewidth]{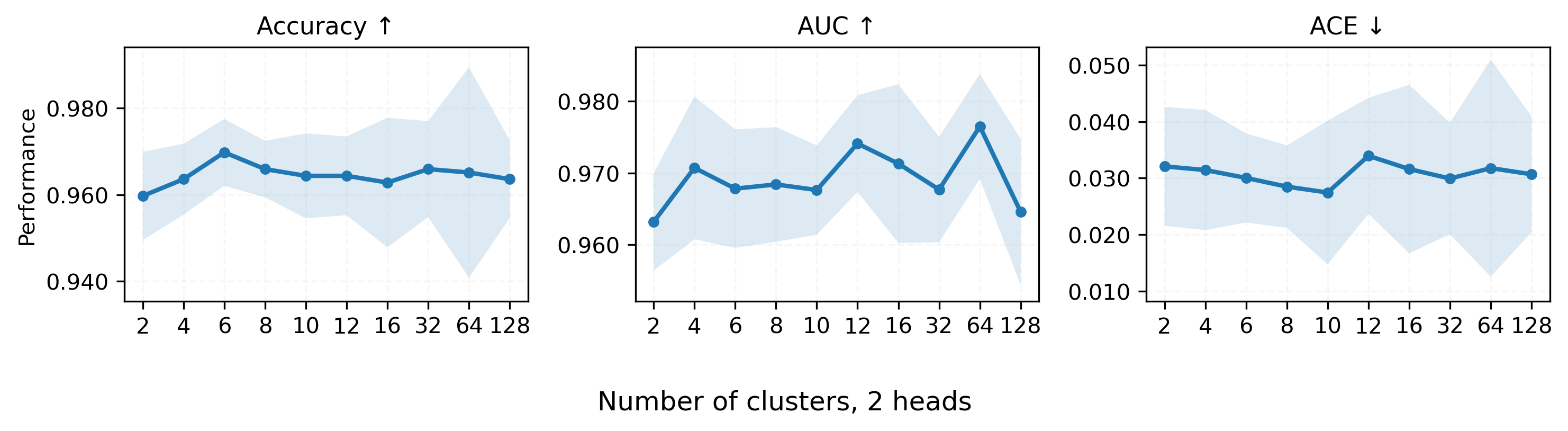}
\vspace{0.1em}

\includegraphics[width=\linewidth]{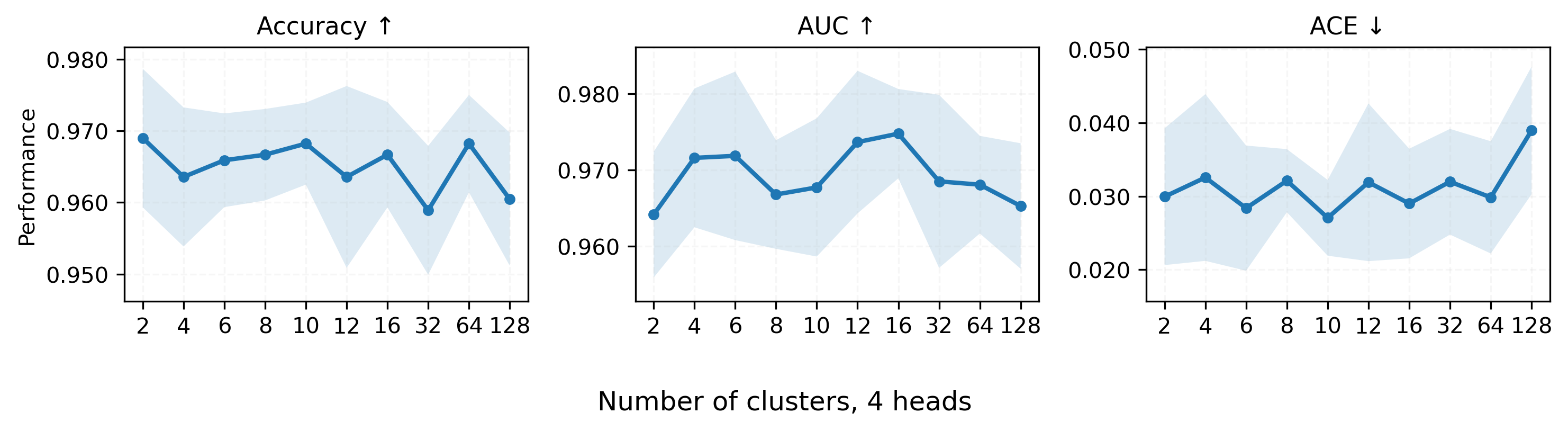}
\vspace{0.1em}

\includegraphics[width=\linewidth]{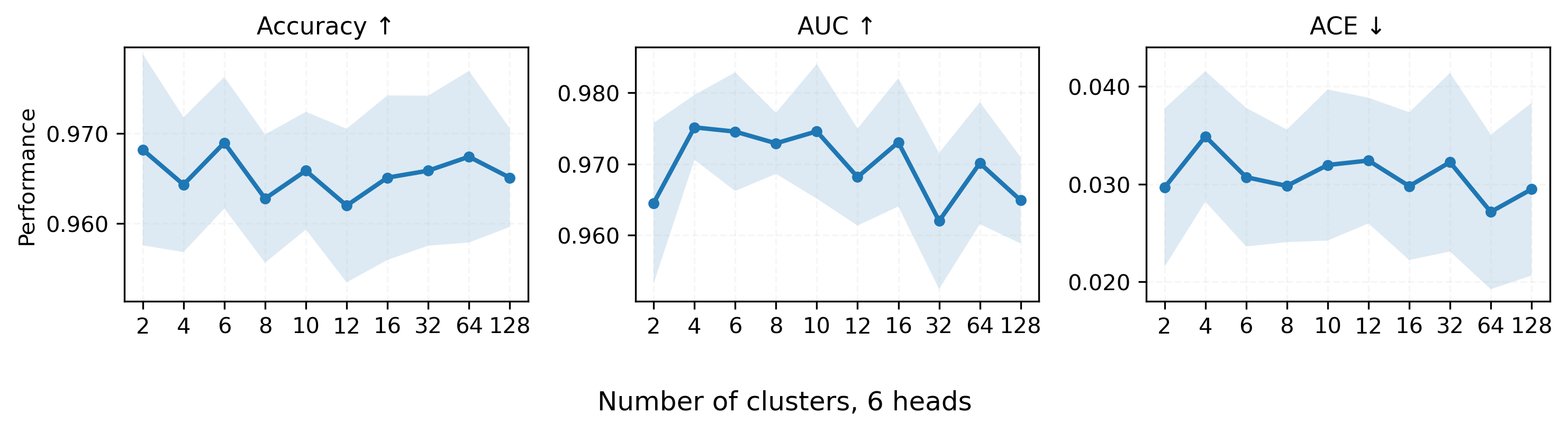}
\vspace{0.1em}

\includegraphics[width=\linewidth]{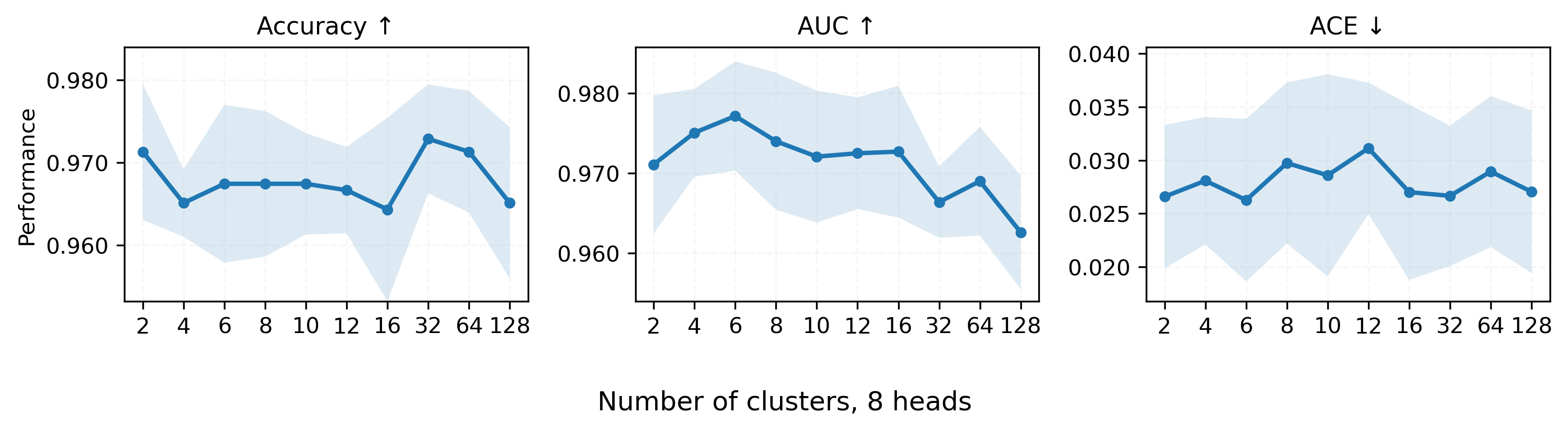}
\vspace{0.1em}

\includegraphics[width=\linewidth]{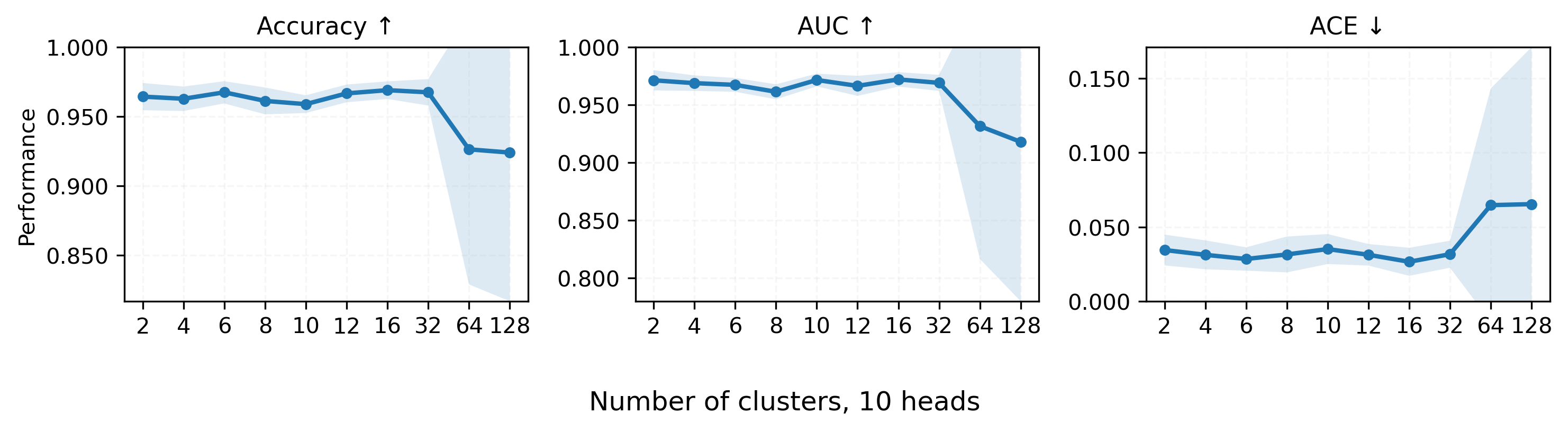}
\vspace{0.1em}

\includegraphics[width=\linewidth]{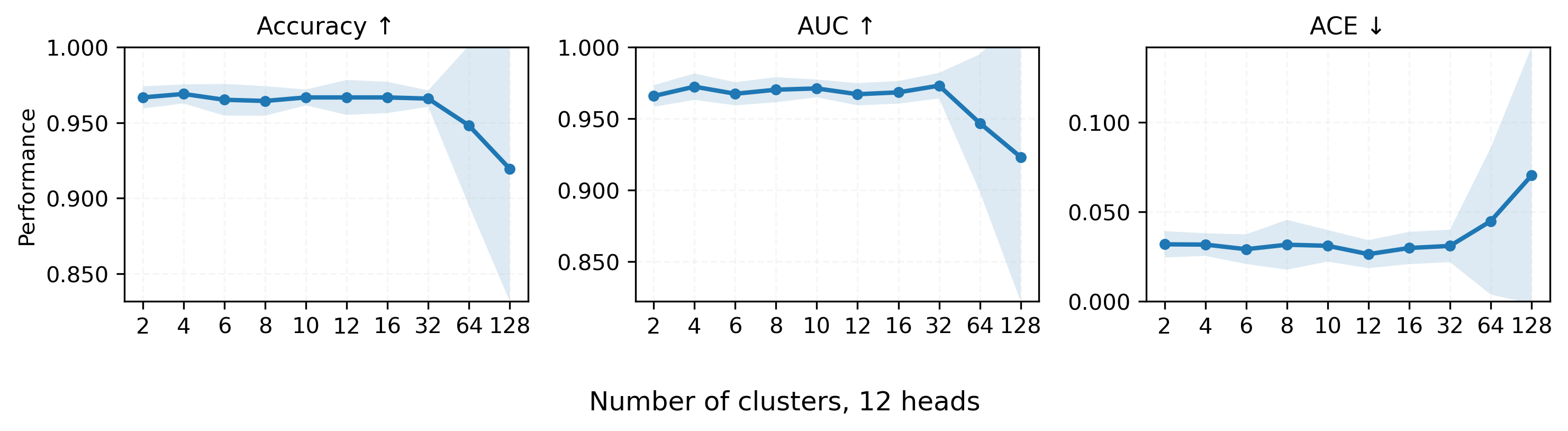}

\end{minipage}}
\caption{
Ablation on the number of clusters for CAMELYON16.
For each row, the number of attention heads is fixed while the number of clusters is varied.
From top to bottom: 1, 2, 4, 6, 8, 10 and 12 heads.
}
\label{fig:ablation_clusters_cam16}
\end{figure}

\begin{figure}[p]
\centering

\resizebox{0.68\textwidth}{!}{%
\begin{minipage}{\textwidth}

\includegraphics[width=\linewidth]{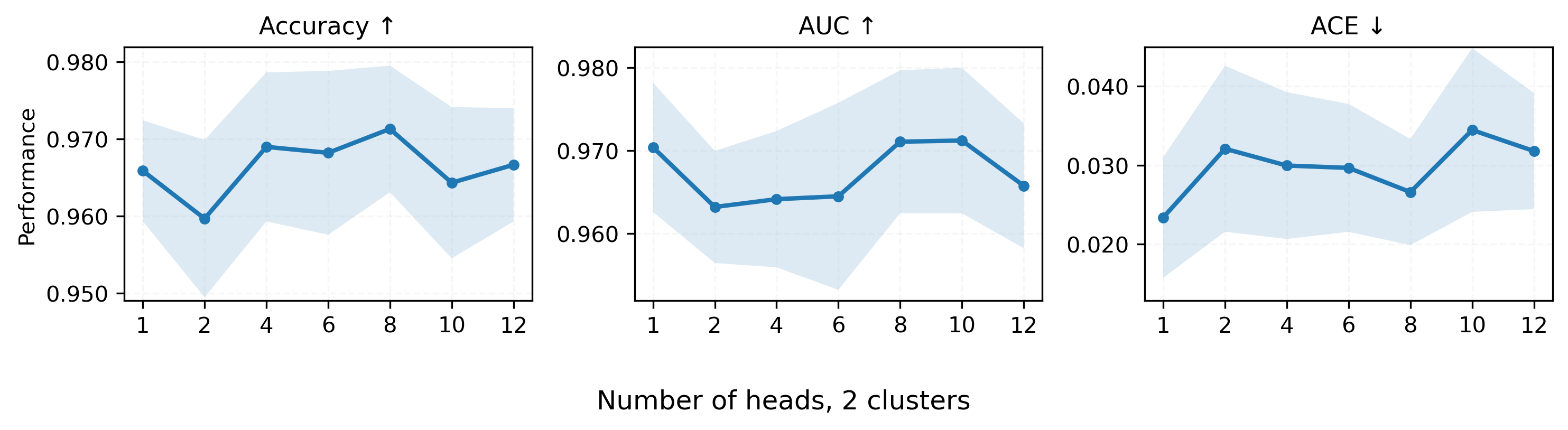}
\vspace{0.1em}

\includegraphics[width=\linewidth]{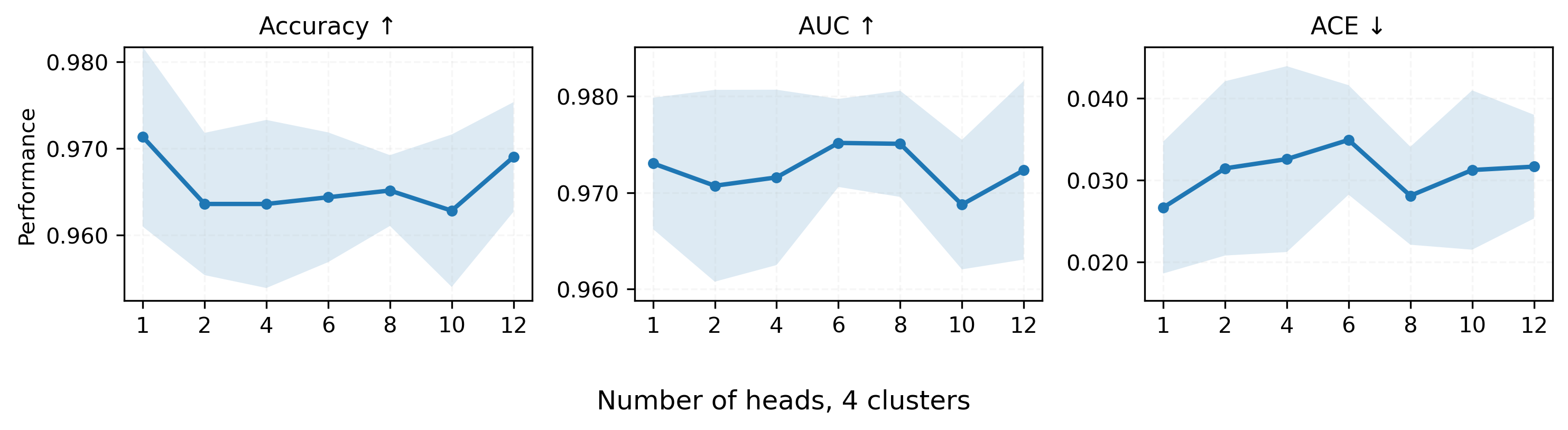}
\vspace{0.1em}

\includegraphics[width=\linewidth]{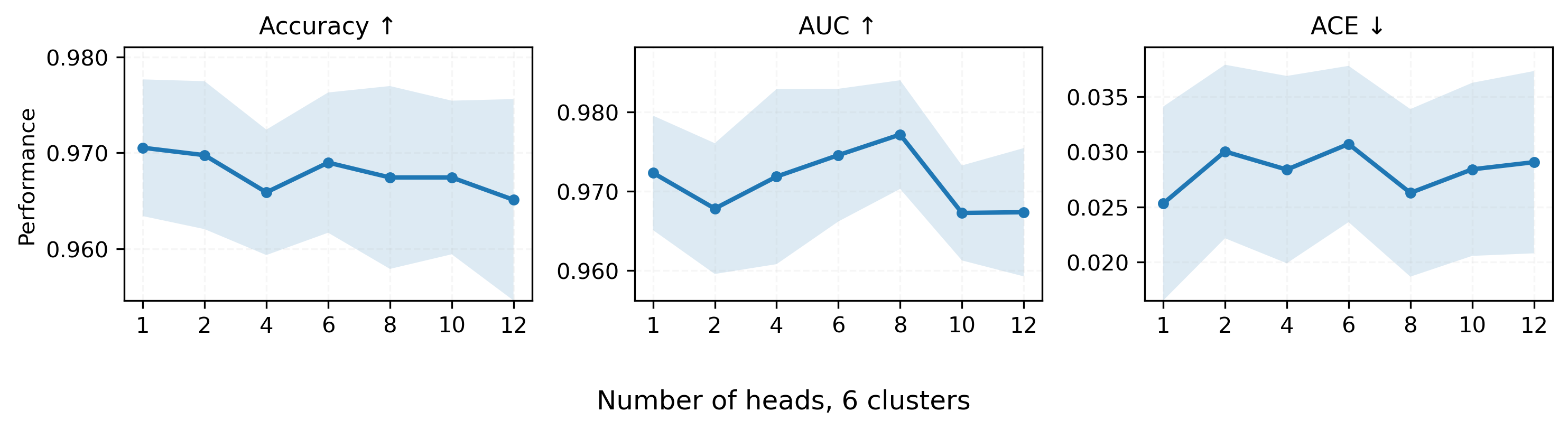}
\vspace{0.1em}

\includegraphics[width=\linewidth]{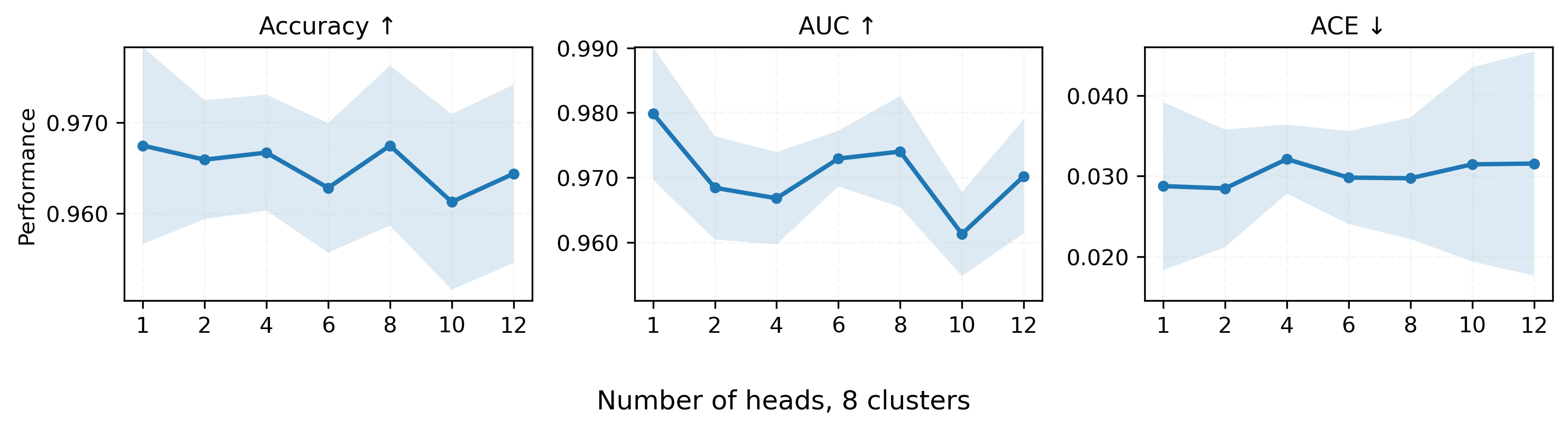}
\vspace{0.1em}

\includegraphics[width=\linewidth]{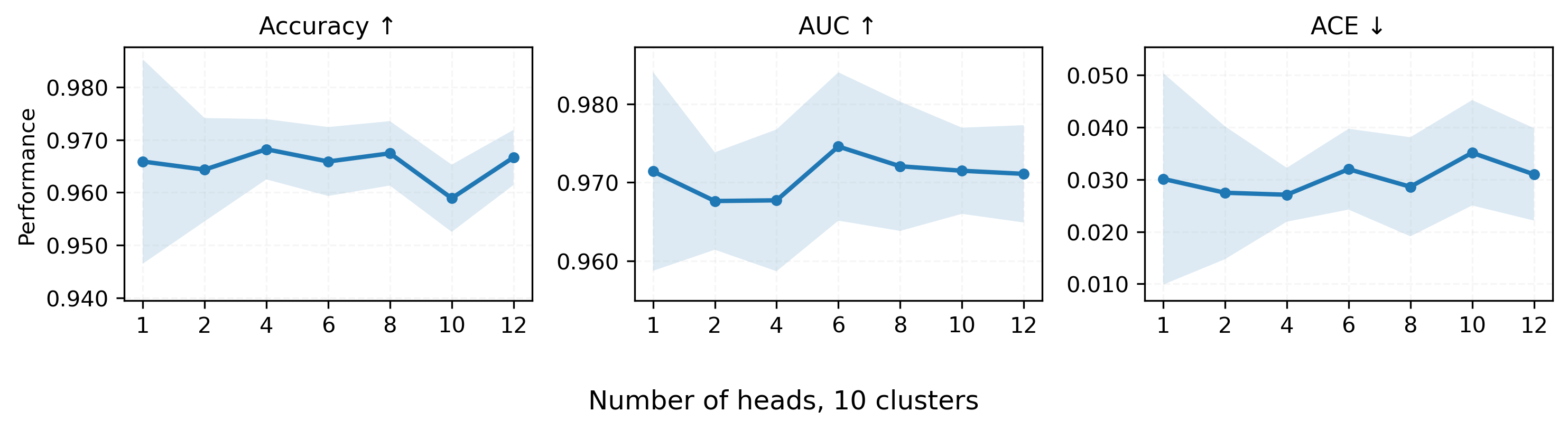}
\vspace{0.1em}

\includegraphics[width=\linewidth]{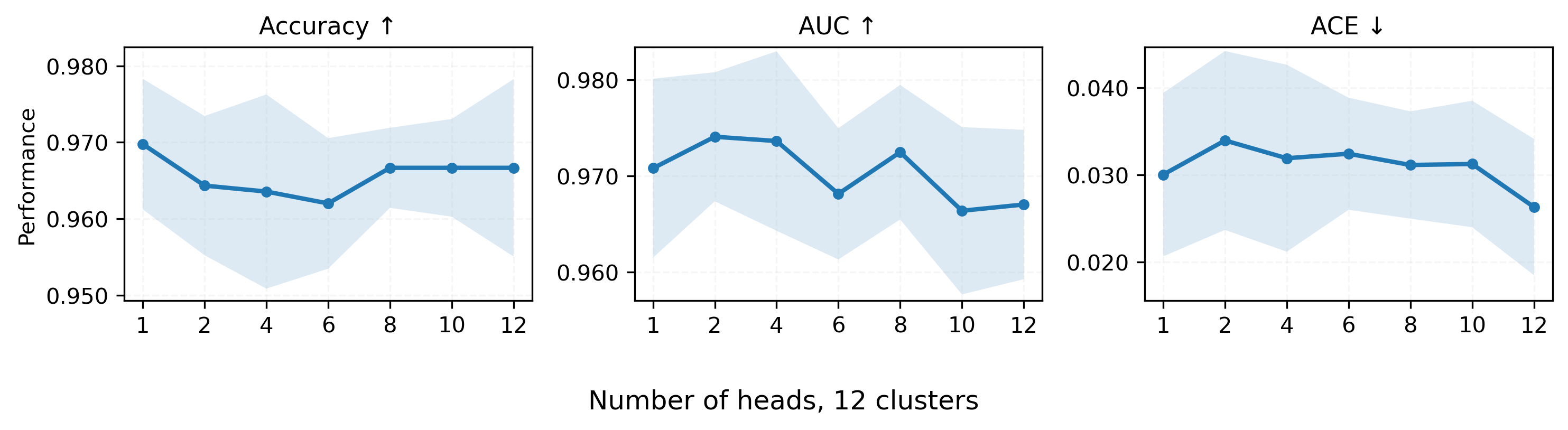}

\end{minipage}}
\caption{
Ablation on the number of attention heads for a single \ours block on CAMELYON16.
For each row, the number of clusters is fixed while the number of heads is varied.
From top to bottom: 2, 4, 6, 8, 10, and 12 clusters.
}
\label{fig:ablation_heads_cam16_0}
\end{figure}

\begin{figure}[p]
\centering

\resizebox{0.9\textwidth}{!}{%
\begin{minipage}{\textwidth}

\includegraphics[width=\linewidth]{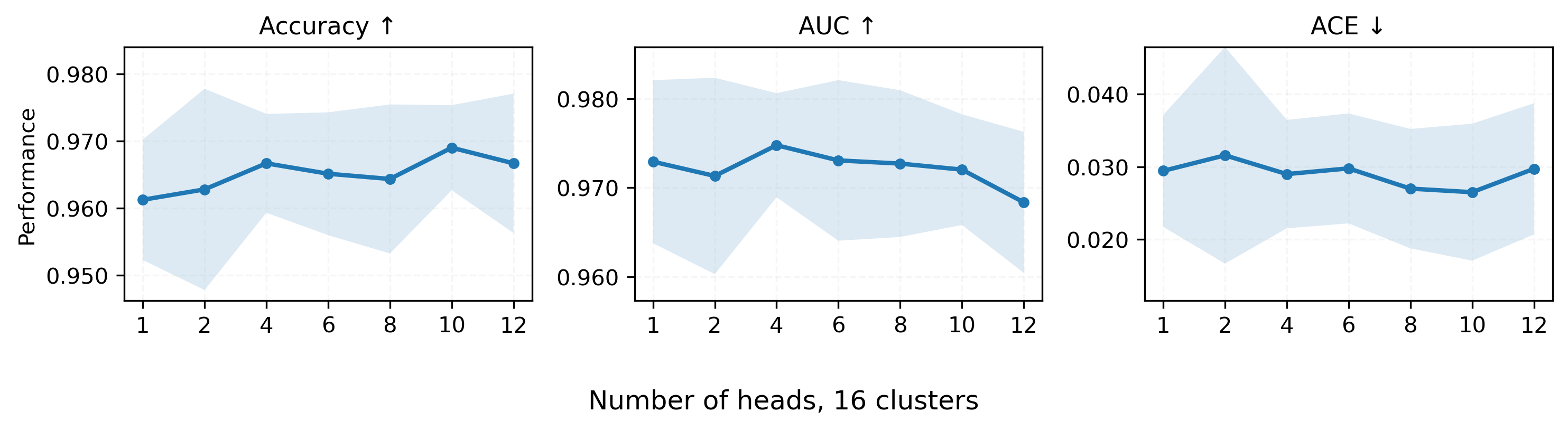}
\vspace{0.1em}

\includegraphics[width=\linewidth]{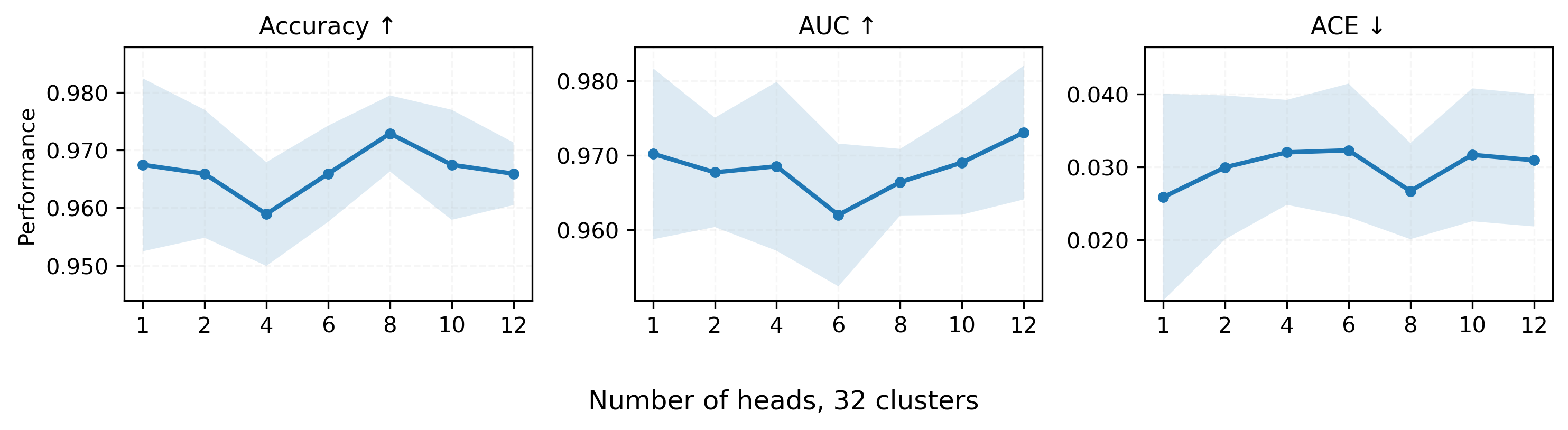}
\vspace{0.1em}

\includegraphics[width=\linewidth]{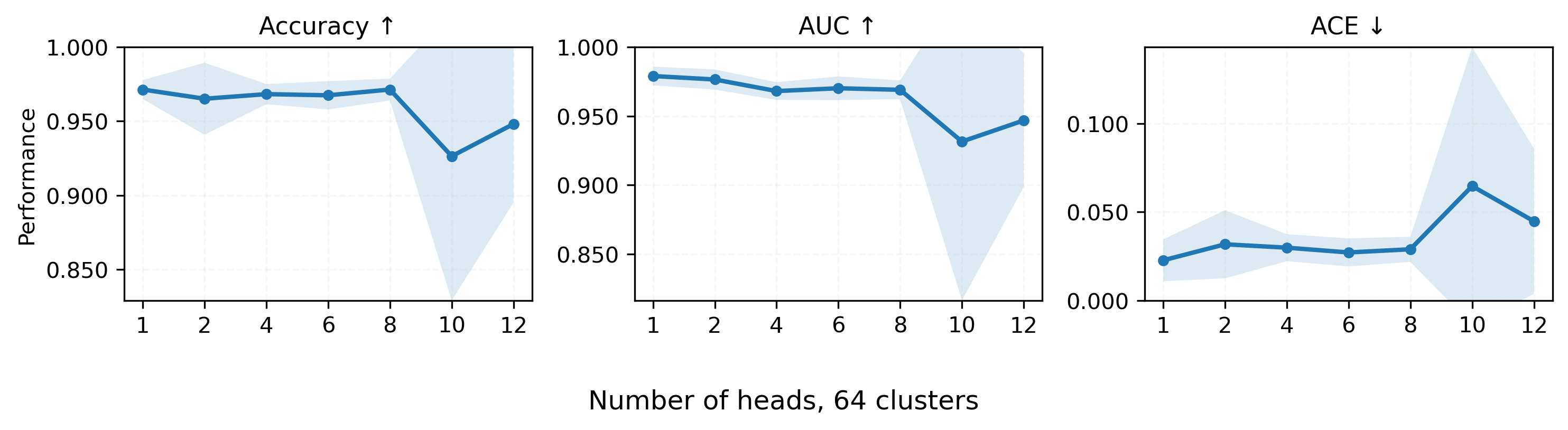}
\vspace{0.1em}

\includegraphics[width=\linewidth]{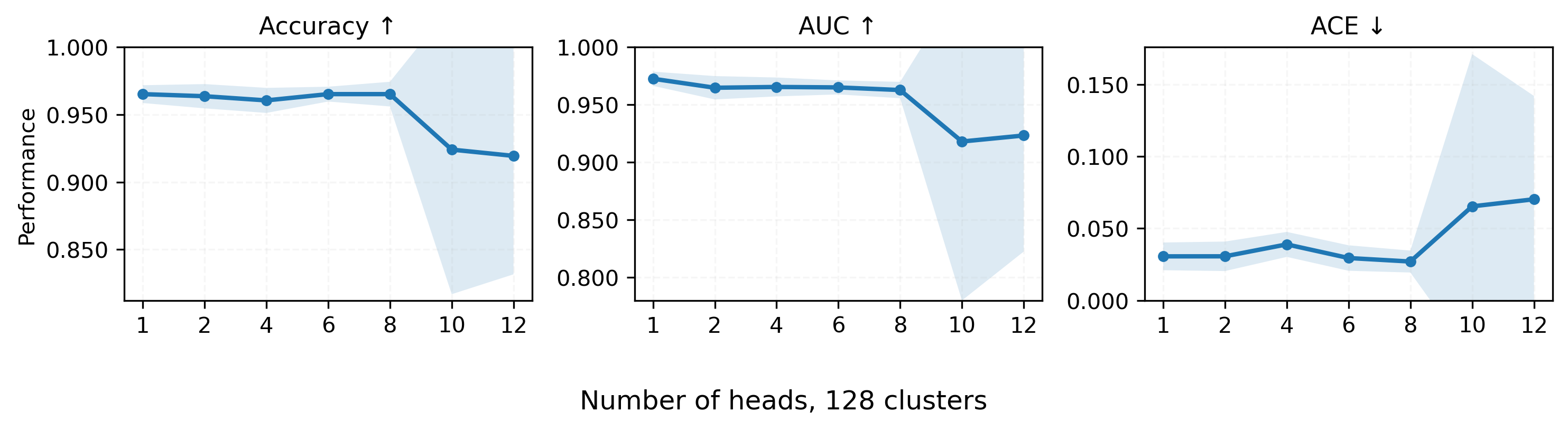}

\end{minipage}}
\caption{
Ablation on the number of attention heads for a single \ours block on CAMELYON16.
For each row, the number of clusters is fixed while the number of heads is varied.
From top to bottom: 16, 32, 64, and 128 clusters.
}
\label{fig:ablation_heads_cam16_1}
\end{figure}

\begin{figure}
    \centering
    \includegraphics[width=0.75\linewidth]{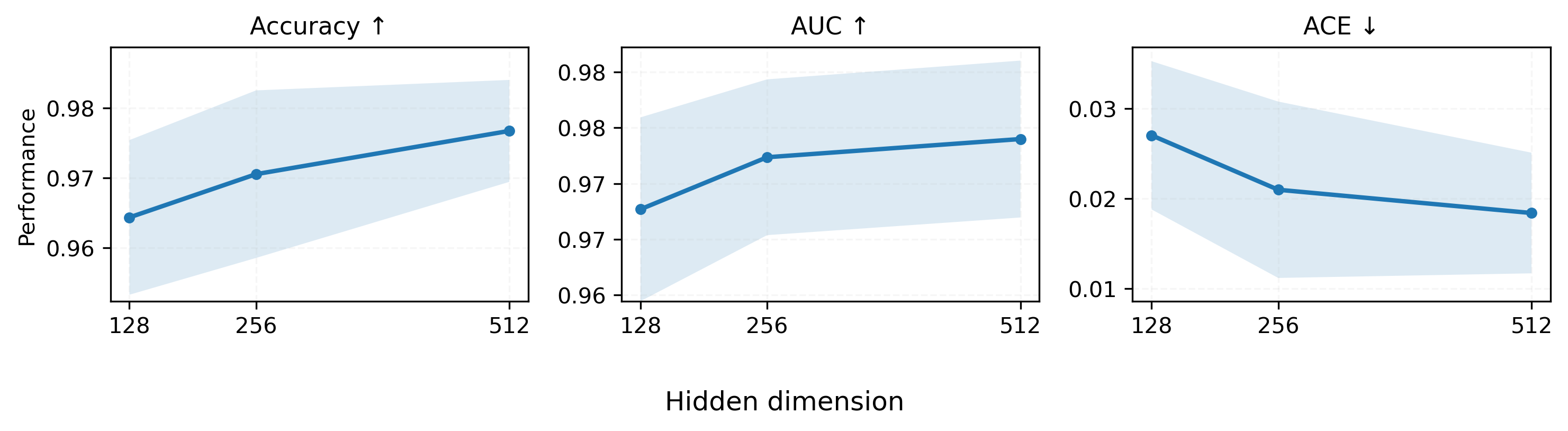}
\caption{
    Ablation on the dimensionality of the input projection layer on CAMELYON16.
    We vary the number of hidden units while keeping the number of clusters (16)
    and attention heads (8) fixed.
}
    \label{fig:ablation_hdims_cam16}
\end{figure}

\begin{figure}
    \centering
    \includegraphics[width=0.75\linewidth]{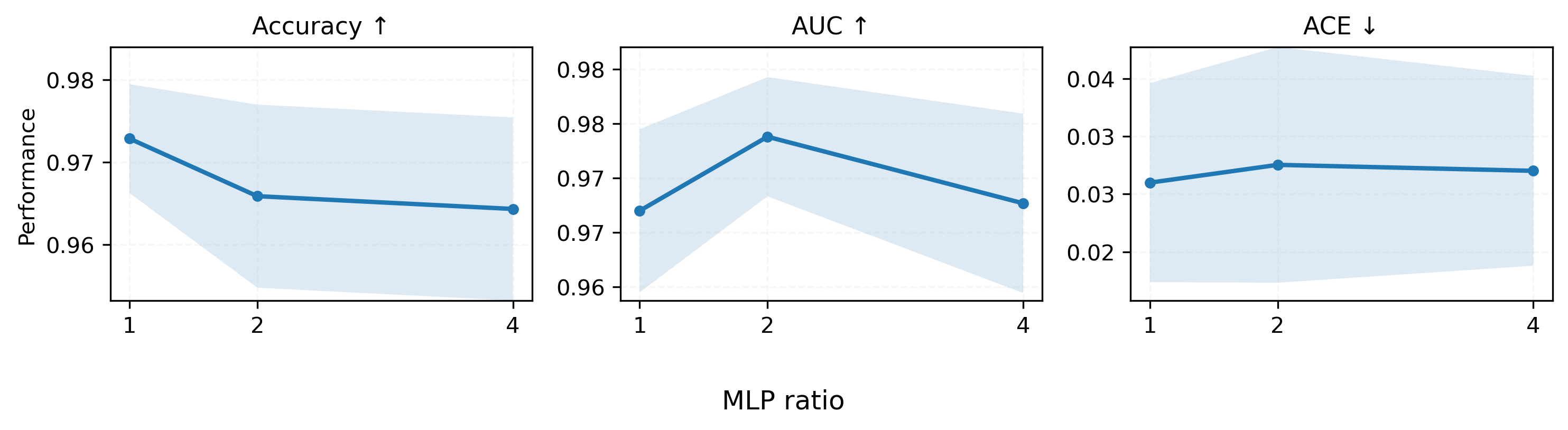}
\caption{
    Ablation on the MLP expansion ratio in the \ours block on CAMELYON16.
    We vary the expansion factor of the feed-forward network
    (MLP ratio $\in \{1, 2, 4\}$) while keeping all other components fixed (16 clusters, 8 heads).
}
    \label{fig:ablation_mlp_cam16}
\end{figure}

We ablate key architectural choices of the Transolver block along three orthogonal axes: the number of clusters used for tokenization ($M$), the number of attention heads ($H$), and the MLP expansion ratio. Experiments are conducted on CAMELYON16, with full sweeps reported in Figures A~\ref{fig:ablation_clusters_cam16}, \ref{fig:ablation_heads_cam16_0}, \ref{fig:ablation_heads_cam16_1}, \ref{fig:ablation_hdims_cam16} and \ref{fig:ablation_mlp_cam16}.


\begin{table}[h]
    \centering
\begin{tabular}{lcccc}
\toprule
Model & Training (s) & Inference (s) & Params (M) & FLOPs (G) \\
\midrule
ABMIL~\cite{ilse_attention-based_2018}      & $\mathbf{5.5}$ & $\mathbf{0.8}$ & $0.660$ & $1.31$ \\
CLAM~\cite{lu_data-efficient_2021}       & $7.0$         & $0.9$         & $0.920$ & $1.84$ \\
TransMIL~\cite{shao_transmil_2021}   & $13.4$        & $1.2$         & $2.67$  & $85.02$ \\
DGRMIL~\cite{zhu_dgr-mil_2025}    & $16.7$        & $1.5$         & $4.34$  & $79.88$ \\
BayesMIL~\cite{cui_bayes-mil_2023}& $9.5$         & $1.1$         & $1.32$  & $2.63$ \\
SGPMIL~\cite{lolos_sgpmil_2025}    & $9.0$         & $1.0$         & $1.21$  & $2.43$ \\
\ours & $6.3$     & $\mathbf{0.8}$         & $\mathbf{0.314}$ & $\mathbf{0.628}$ \\
\bottomrule
\end{tabular}

    \caption{Training and inference times (in seconds) and model sizes (number of trainable parameters in millions, M). 
    Training times are averaged over 30 epochs, while inference times correspond to processing the full test set of $129$ slides.}
    \label{tab:timing_comparison}
\end{table}



\clearpage


\clearpage

\clearpage
\clearpage


\clearpage

\end{document}